\documentclass[lettersize,journal]{IEEEtran}
\usepackage{amsmath,amsfonts}
\usepackage{algorithmic}
\usepackage{algorithm}
\usepackage{array}
\usepackage[caption=false,font=normalsize,labelfont=sf,textfont=sf]{subfig}
\usepackage{textcomp}
\usepackage{stfloats}
\usepackage{url}
\usepackage{verbatim}
\usepackage{graphicx}
\usepackage{cite}
\usepackage{xcolor} 
\usepackage{booktabs}
\usepackage{multirow}
\usepackage{adjustbox}
\usepackage{amsmath, amssymb, amsthm}

\newcommand{\shortname}{\textit{Stratify}}

\begin{document}

\title{Stratify: Rethinking Federated Learning for Non‑IID Data through Balanced Sampling}

\author{Hui Yeok Wong\textsuperscript{1}, Chee Kau Lim\textsuperscript{1,*}, Chee Seng Chan\textsuperscript{1}\\ \textsuperscript{1}Department of Artificial Intelligence, Universiti Malaya, Malaysia
\thanks{\textsuperscript{*}Corresponding Author: limck@um.edu.my\\Preprint under review.}}

\maketitle

\begin{abstract}
Federated Learning (FL) on non-independently and identically distributed (non-IID) data remains a critical challenge, as existing approaches struggle with severe data heterogeneity. Current methods primarily address symptoms of non-IID by applying incremental adjustments to Federated Averaging (FedAvg), rather than directly resolving its inherent design limitations. Consequently, performance significantly deteriorates under highly heterogeneous conditions, as the fundamental issue of imbalanced exposure to diverse class and feature distributions remains unresolved. This paper introduces \shortname, a novel FL framework designed to systematically manage class and feature distributions throughout training, effectively tackling the root cause of non-IID challenges. Inspired by classical stratified sampling, our approach employs a Stratified Label Schedule (SLS) to ensure balanced exposure across labels, significantly reducing bias and variance in aggregated gradients. Complementing SLS, we propose a label-aware client selection strategy, restricting participation exclusively to clients possessing data relevant to scheduled labels. Additionally, \shortname\ incorporates a fine-grained, high-frequency update scheme, accelerating convergence and further mitigating data heterogeneity. To uphold privacy, we implement a secure client selection protocol leveraging homomorphic encryption, enabling precise global label statistics without disclosing sensitive client information. Extensive evaluations on MNIST, CIFAR-10, CIFAR-100, Tiny-ImageNet, COVTYPE, PACS, and Digits-DG demonstrate that \shortname\ attains performance comparable to IID baselines, accelerates convergence, and reduces client-side computation compared to state-of-the-art methods, underscoring its practical effectiveness in realistic federated learning scenarios.
\end{abstract}

\begin{IEEEkeywords}
Federated Learning, Non-IID Data, Label Skew, Feature Skew.
\end{IEEEkeywords}

\section{Introduction}

\IEEEPARstart{F}{ederated} learning (FL) has emerged as a promising paradigm for distributed machine learning that preserves data privacy by enabling clients to collaboratively train a global model without sharing their raw data \cite{guerra2023cost}. However, a persistent challenge in FL is that client data are often non-independently and identically distributed (non-IID). In practical settings, clients typically possess heterogeneous datasets with imbalanced class distributions or varying feature characteristics. Such data heterogeneity leads to imbalanced local updates, which, when aggregated, bias the global model and slow down convergence. 

Existing research has developed various algorithms to address non-IID challenge. The common approaches used include: i) cluster similar clients for collaboration \cite{Briggs2020, jothimurugesan2023federated, Vahidian2023, Xiao2021}, ii) weighted local model parameters aggregation \cite{Huang2021, Luo2022}, iii) data augmentation \cite{Chen2023, li2024feature}, iv) apply correction term to restrict local deviation from global model \cite{Li2020, Xu2022, Karimireddy2020}, and v) ensemble-based approach \cite{zhao2023ensemble, Shi2023}. These approaches primarily focus on avoiding collaboration among non-IID clients to align with FedAvg's IID assumption \cite{mcmahan2017communication} or applying patches on FedAvg to deal with diverged local updates. However, such methods address only the symptoms rather than directly targeting the root cause: the uneven exposure of the global model to diverse label and feature distributions.


In this work, we introduce \shortname, a novel framework for federated learning that addresses the non-IID challenge by rethinking the training process. Our method is built on the concept of a Stratified Label Schedule (SLS), which draws inspiration from classical stratified sampling \cite{thompson2012sampling}. By constructing a balanced and randomized schedule of labels based on controlled repetition frequencies, \shortname\ ensures that the global model is systematically exposed to all classes, thereby mitigating bias and reducing the variance of the aggregated gradients. Our theoretical analysis, grounded in the principles of stratified sampling \cite{thompson2012sampling} and variance decomposition \cite{hamilton2020time} demonstrates that this balanced update scheme yields an unbiased gradient estimator with improved stability.

Complementing the SLS, our framework employs a label-aware client selection mechanism that restricts participation to those clients possessing the relevant data for the designated label. Additionally, we propose a novel learning strategy, in which the global model is sequentially updated using individual samples or small batch samples. This fine-grained update process accelerates convergence and further alleviates the adverse effects of data heterogeneity. To ensure that these guided updates are implemented in a privacy-preserving manner, we develop a secure client selection protocol that leverages a masked label representation based on the Cheon-Kim-Kim-Song (CKKS) homomorphic encryption (HE) scheme. This approach enables accurate global label statistics to be computed without exposing clients’ sensitive information.

Our contributions are fourfold.
\begin{itemize}
    \item \shortname\ introduces SLS and corresponding label-aware client selection to achieve balanced and robust model updates under non-IID conditions.
    \item Our novel learning strategy provides high-frequency, fine-grained updates that lead to faster convergence.
    \item Our secure implementation, which utilizes masked label representation with CKKS, ensures that these benefits are realized without compromising client privacy.
    \item Extensive experiments on benchmark datasets (i.e.,~MNIST, CIFAR-10, CIFAR-100, Tiny-ImageNet, COVTYPE, PACS, and Digits-DG) demonstrate that \shortname\ achieves performance close to the IID baseline, converges in fewer rounds, and achieves lower overall local computation workload per client, compared to state-of-the-art methods.
\end{itemize}  

\section{Related Work}
\subsection{Client Clustering}
Various algorithms have been developed to address non-IID data challenge in FL. The clustering approach
groups clients with similar data using similarity measure to enable collaboration among homogeneous clients. Briggs et al. \cite{Briggs2020} and CFMTL \cite{Xiao2021} use local update parameters with a distance metric to find similar clients. FEDCOLLAB \cite{Bao2023} compares client data distributions by training a global discriminator where a balanced accuracy around 50\% indicates similar client data. FedDrift \cite{jothimurugesan2023federated} measures the maximum loss degradation when applying local models to each other's data. Clients with loss values below a threshold are merged. PACFL \cite{Vahidian2023} uses truncated SVD to extract key features of each client's data, and clusters clients with similar data based on principal angles spanned by these vectors. However, clustering approach is ineffective for small participating clients and each holds a diverse data. The likelihood of finding clients with similar data can be slim as the similarity measures used in these research identify homogeneous clients based on their overall datasets \cite{Luo2022}. Furthermore, these methods have to maintain multiple global models which increases complexity and limits generalization across clients. 

\subsection{Weighted Aggregation}
Some works tackled non-IID issue by minimizing the variance of client updates through weighted contributions. FedAMP \cite{Huang2021} and APPLE \cite{Luo2022} maintains a personalized model for each client on server, and update each model through weighted combination of model parameters from other clients. FedAMP measures each pairwise client parameters using Cosine similarity and assigns higher weight to models that are similar during aggregation. APPLE uses the directed relationship (DR) matrix to quantify the influence of other client models on a client model. Each client DR matrix is adaptively updated during training, with weights adjusted based on the gradients of the loss function. If other client model negatively impacts the client, the corresponding weight in the DR matrix moves closer to zero or becomes negative to minimize harmful influence. However, this approach faces the same limitations as clustering approach as it relies on similarity metric to guide collaboration. In highly diverse client datasets, this approach will struggle to assign meaningful weights. 

\subsection{Local Model Regularization}
FedProx \cite{Li2020}, SCAFFOLD \cite{Karimireddy2020}, and DOMO \cite{Xu2022} modify the aggregation process to stabilize learning on non-IID data. FedProx adds a proximal regularization on the local model against the global model to keep the updated parameter close to the global model. SCAFFOLD estimates the difference in update directions for each client and the global model. This difference captures the drift of the client parameters from the global model, which is then used to correct the local update. DOMO adjusts the local model momentum of the client to the direction of global momentum. However, in cases of extreme data heterogeneity, the differences between local and global model can be too large, making regularization alone insufficient to bridge the divergence effectively. 

\subsection{Data Augmentation}
Data augmentation methods in FL aim to mitigate non-IID effects by enhancing local data diversity. FRAug \cite{Chen2023} tackles feature skew by training a shared generator to produce client-agnostic embeddings. These embeddings are transformed by a local Representation Transformation Network (RTNet) into client-specific residuals, which are then used to augment real feature representations to enhance model robustness against feature distribution shifts. Li et al. \cite{li2024feature} generates synthetic data by extracting essential class-relevant features using class activation map. Before data generation, real features are adjusted along the hard transformation direction. This ensures that the synthetic data contains less detailed information and becomes less similar to the real data. The synthetic data is then collect by server and share with clients to combine with local real samples. However, while data augmentation enhances data diversity, it may not fully capture the complexities and nuances of real data, especially when synthetic data must avoid closely mimics the real samples to protect sensitive patterns. Therefore, a method that can directly utilize the diverse real data of clients for training is more practical.

\subsection{Local Model Ensemble}
The Ensemble approach combines the predictions from multiple specialized models to mitigate non-IID and enhance generalization ability across diverse client distributions. Zhao et al. \cite{zhao2023ensemble} groups clients with similar initial gradients into clusters, each collaboratively train a global model. During inference, the predictions from these individual models are combined using weighted coefficients to generate the final prediction. Fed-ensemble \cite{Shi2023} trains an ensemble of $K$ models. For each round, each client receives one of the $K$ models for local training and send the updated model to server for aggregation. After training, all $K$ models are sent to clients and prediction on new data is obtained by averaging the prediction output of all models. However, this approach requires client to hold multiple models locally, which is not storage efficient for resource-constrained devices.

\subsection{Client Selection}
Client selection methods often use clustering to group similar clients, but instead of collaborating within cluster, they sample representative clients from different clusters to enhance model generalization. Fraboni et al. \cite{fraboni2021clustered}, cluster similar clients using their sample size or representative gradients which indicates the difference between their local updates and the global model. Clients are sampled from these different clusters for collaboration, ensuring diverse representation during aggregation. FedSTS \cite{gao2024fedsts} uses compressed gradient to filter out redundant information in the raw gradients for better grouping clients into strata. During client selection, clients within each stratum with the higher gradient norms are given higher probabilities of being selected, ensuring that the most significant updates are included. However, in cases of extreme data diversity among clients, these methods will fail as a client does not clearly belong to a single cluster due to diverse data characteristics. This results in poorly defined groups that lead to a suboptimal client selection and ultimately degrading the model performance. 

Despite extensive research on improving FL for non-IID data, existing approaches still face limitations including reliance on potential noisy similarity metrics for clustering, and the increased complexity and storage inefficiency from maintaining multiple models. In contrast, our proposed solution effectively learn on non-IID client data without being restricted by these limitations. We present \shortname\ in the next section.

\begin{figure*}[!t]
\centering
\hspace{-18pt}
\subfloat[Single-sample learning]{\includegraphics[width=9cm, height=7.5cm]{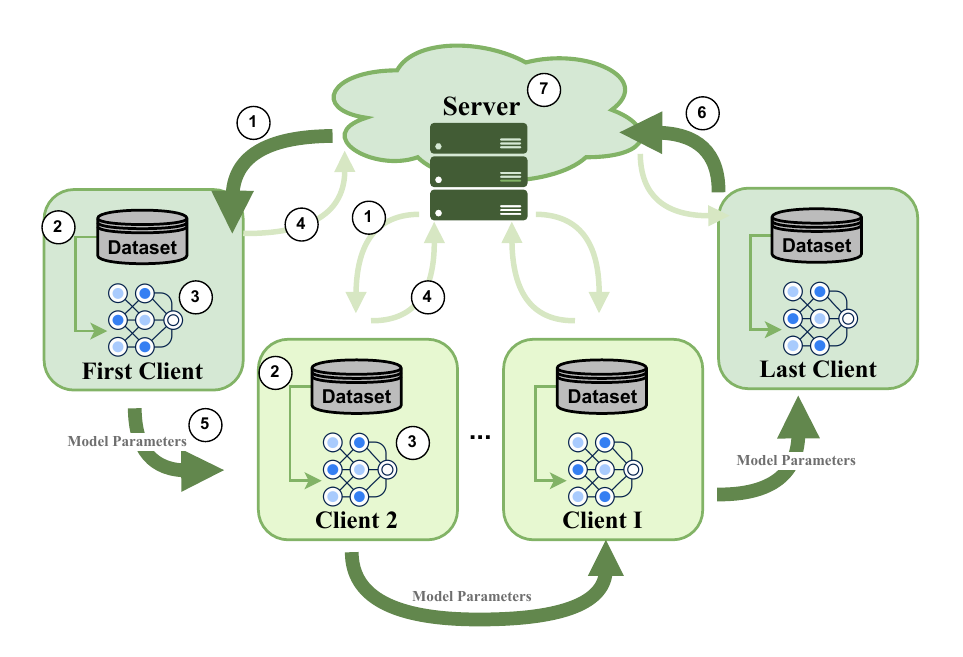}%
\label{fig_sl}}
\hfil
\subfloat[Batch-data learning]{\includegraphics[width=8.5cm, height=7cm]{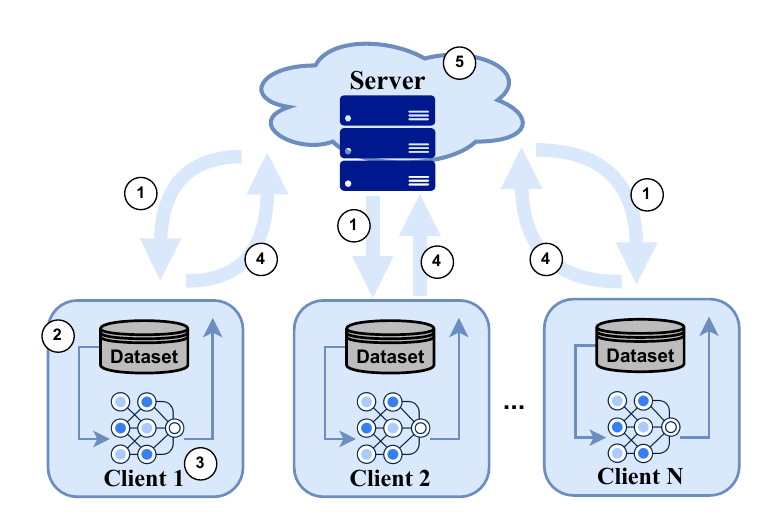}%
\label{fig_bl}}
\caption{Overview of \shortname\ Training Process. (a) Single-sample learning: (a1) Send initialized global model parameter (only first client), masked labels to train, and next client address to current client, (a2) Convert masked label to real label, (a3) Update model sequentially on the required labels, (a4) Signal server to send next masked labels to train to next client, (a5) Send updated model to next client, Repeat step a1 to a5 until all masked labels to train are pulled from $SLS$, (a6) Send the final updated model to server once the last client completes its training and (a7) Broadcast the received global model to all clients. (b) Batch-data learning: (b1) Send masked labels to train to all selected clients for the current batch, (b2) Convert masked label to real label, (b3) Compute summed gradient, (b4) Return summed gradient, and (b5) Compute a unified gradient to update global model by summing up the clients' summed gradients and then dividing it with the batch size, and send the updated model to all selected clients in next batch. Repeat step b1 to b5 until all masked labels to train are pulled from $SLS$.} 
\label{fig_sim}
\hspace{-18pt}
\end{figure*}

\section{\shortname\ Framework}

Federated learning on non-IID data often leads to imbalanced local updates and biased global models when using conventional aggregation techniques (e.g., FedAvg). To address these challenges, our proposed \shortname\ framework introduces a guided training mechanism that ensures balanced exposure to all class labels during global model updates. Central to our approach is the Stratified Label Schedule (SLS), a mechanism inspired by classical stratified sampling that produces unbiased gradient estimates with reduced variance. The following subsections detail the construction of the SLS, its theoretical foundation, and how it informs both client selection and model training.

\subsection{Stratified Label Schedule}
\label{SLS-A1}

To mitigate the adverse effects of data heterogeneity, the server aggregates the unique labels from all participating clients and constructs a balanced schedule by repeating each label $l$ a controlled number of times according to $f_l$. We define the Stratified Label Schedule (SLS) as
\[ \text{SLS} = \text{Shuffle}\!\left(\bigcup_{l \in \mathcal{L}} f_l\{l\}\right),\]
\noindent where $\mathcal{L}$ is the set of all unique labels, $f_l\{l\}$ denotes that label $l$ is repeated $f_l$ times, and $f_l$ can be set uniformly or proportionally to a capped global count for each label $l$ to prevent overrepresentation.

Shuffling randomizes the order so that each training cycle exposes the global model to a diverse and balanced mix of classes. In each training iteration, the server selects the next label $l$ from the SLS and assigns training tasks exclusively to those clients that possess samples with label $l$.


\subsection{Theoretical Analysis: Stratified Sampling and Variance Reduction}
\label{SLS-A2}

Inspired by classical stratified sampling where a population is divided into homogeneous subgroups (strata) and samples are drawn from each to reduce estimator variance. We view the SLS as a mechanism for enforcing balanced label representation. In our context, each label $l$ forms a stratum. Let $g_l$ denote the gradient computed from local sample with label $l$. The aggregated gradient update is then computed as
\[ \hat{g} = \frac{1}{|\text{SLS}|} \sum_{l \in \text{SLS}} g_l,\]

By the linearity of expectation, we have
\[ \mathbb{E}[\hat{g}] = \sum_{l \in \mathcal{L}} P(l)\,\mathbb{E}[g_l],\]
\noindent where the probability-proportional-to-size for label $l$ is defined as:
\[ P(l) = \frac{f_l}{\sum_{l\in \mathcal{L}} f_{l}},\]
\noindent and with $f_l$ denoting the frequency of label $l$ in SLS. By enforcing a uniform frequency across all labels $l \in \mathcal{L}$, $P(l)$ becomes uniform. As a result, each label contributes equally to the aggregated gradient. This confirms that SLS yields an unbiased and balanced gradient estimator by constructing an equal representation across all labels. 

Furthermore, by the law of total variance:
\[ \operatorname{Var}(\hat{g}) = \mathbb{E}\left[\operatorname{Var}(\hat{g}\mid l)\right] + \operatorname{Var}\left(\mathbb{E}[\hat{g}\mid l]\right).\]
\noindent The term $\operatorname{Var}\left(\mathbb{E}[\hat{g}\mid l]\right)$ reflects the variance due to imbalanced contributions across labels. Enforcing balanced sampling via the SLS minimizes this term relative to naive sampling, leading to reduced overall variance and more stable convergence in non-IID settings. 

\subsection{Client Selection Based on the SLS}
\label{SLS-A3}

\begin{algorithm}[t]
    \small
    \caption{Single-Sample Learning Client Selection.}
    \label{alg:sscs}
    \begin{algorithmic}[1] 
        \STATE $C_{select} \gets [\:]$
        \FOR{$p \in Placeholder_{unique}$}
            \STATE $availClients_p \gets$ \textbf{Update} client pool by retaining clients 
            \STATE \hspace{2.1cm} using uniform or weighted client selection
        \ENDFOR
        \FOR{$(i, \:p)$ in enum($P_{chunk}$)}
            \FOR{$c_j \in availClients_p $}
                \STATE $next\_i \gets i$
                \WHILE{True}
                    \STATE $next\_i \gets \min(next\_i + 1,\:chunkSize - 1)$
                    \IF{$c_j \in availClients_p \textbf{ of } P_{chunk}[next\_i]$}
                        \STATE $conseqLen_{c_j} \gets conseqLen_{c_j} + 1$
                        \IF{$next\_i = chunkSize - 1$}
                            \STATE \text{\textbf{break} out from while loop}
                        \ENDIF
                    \ELSE
                        \STATE \text{\textbf{break} out from while loop}
                    \ENDIF

                \ENDWHILE
            \ENDFOR
            \STATE $C_{select}[i\colon i + conseqLen_{c_j} + 1] \gets \text{\textbf{Select} any one $c_j$}$
            \STATE \hspace{2cm} $\text{with $conseqLen_{c_j} = \max(conseqLen)$}$
            \STATE \text{\textbf{continue} outer for loop at $i \gets i + conseqLen_{c_j} + 1$}
        \ENDFOR
    \end{algorithmic}
\end{algorithm}

Building on the SLS mechanism, the server employs a label-aware client selection strategy. For each training iteration, after extracting the next label $l$ from the SLS, the server selects only those clients whose local datasets contain samples with label $l$. This targeted selection ensures that the gradient update for that iteration is computed solely from data corresponding to $l$, thereby preserving the balanced representation and variance reduction benefits established in Section \ref{SLS-A1}. Algorithm \ref{alg:sscs} and \ref{alg:blcs} detail the client selection process. \shortname\ supports two client selection strategies: 

(i) \textbf{Uniform Client Selection:} An equal probability of client selection that maximizes label privacy. This strategy relies only on masked label availability, ensuring strong label privacy without requiring the server to access any label-size-related metadata. Uniform selection works well in most non-IID scenarios, particularly the label skew setting and mild feature skew cases. The actual label identities of clients are masked using abstract placeholders, ensuring that the true semantic label values remain hidden from server. This allows server to perform label-aware selection without compromising label privacy (see Section \ref{Tech-A2}).

(ii) \textbf{Weighted Client Selection:} An adjusted selection probabilities based on the proportion of each client's label count relative to the global label count, computed using HE. This strategy mitigates the risk of overfitting to the clients feature domains with disproportionately large data during the later training iterations, by enforcing a consistent exposure to diverse client domains. While it introduces a slight relaxation in label privacy, the obfuscated label identifiers ensure that the label proportions are untraceable to the actual categories they represent. This trade-off allows for better handling of severe feature heterogeneity, while maintaining adequate label privacy safeguard. 

In Algorithm \ref{alg:sscs}, the $chunkSize$ parameter is introduced to process labels from SLS in chunks, allowing the server to group multiple labels and assign them consecutively to the same client, if available. This parameter also effectively controls the model's exposure frequency to diverse client data between chunks, which is especially useful for feature-skew scenarios. Whereas the $batchSize$ parameter in Algorithm \ref{alg:blcs}, controls the number of labels used in each global batch update.

\begin{algorithm}[t]
    \small
    \caption{Batch-Data Learning Client Selection.}
    \label{alg:blcs}
    \begin{algorithmic}[1] 
        \STATE $C_{select} \gets [\:]$
        \FOR{$p \in$ set($P_{batch}$)}
            \STATE $N_p \gets \text{Freq}(p, P_{batch})$
            \STATE $C^p_{elem} \gets \text{\textbf{Select} $N_p$ number of clients from $availClients_p$}$
            \STATE \hspace{1.5cm}$\text{using uniform or weighted clients selection}$
        \ENDFOR
    \FOR{$(i, \:p)$ in enum($P_{batch}$)}
        \STATE $C_{select}[i] \gets C^p_{elem}[0], \ C^p_{elem} \gets C^p_{elem}[1\colon\!]$
    \ENDFOR

    \end{algorithmic}
\end{algorithm}

\subsection{Model Training Procedure}
\label{SLS-A4}

\begin{algorithm}[t]
    \small
    \caption{Single-Sample Per Iteration Model Training.}
    \label{alg:ssmt}
    \begin{algorithmic}[1] 
        \STATE $\textcolor{red}{\text{\texttt{\#} Server-side}}$
        \STATE $SLS \gets \text{\textbf{Generate} Stratified Label Schedule}$
        \STATE $queue \gets [\:]$
        
        \WHILE{$|SLS|\textbf{ or } |queue| \neq 0$}
            \IF{$|queue| < 2 \textbf{ and } |SLS| \neq 0$}
                \STATE $P_{chunk} \gets SLS[0\!:\!chunkSize],\ SLS \gets [chunkSize\!:]$
                \STATE $C_{select} \gets \text{\textbf{Run} Algo. \ref{alg:sscs} on $P_{chunk}$}$
                \STATE{\text{\textbf{Group} $P_{chunk}$ placeholders by consecutive identical client}}
                \STATE\hspace{0.3cm}{\text{ID length in $C_{select}$}}
                \STATE $queue \gets \text{\textbf{Append} each group as training task containing}$
                \STATE \hspace{1.5cm}$\text{selected client and placeholders to train details}$
            \ENDIF

            \STATE $task \gets \text{\textbf{Pop} first training task in $queue$} $
            \STATE $task \gets \text{\textbf{Add} next client address from $queue[0]$ client detail}$
            \STATE \hspace{1.4cm}$\text{\textbf{if} $|queue| \neq 0$ \textbf{else} server address}$
            \STATE $P_{notTrain},\ P_{exhaust} \gets \text{\textbf{Send} $task$ to client for training}$
            \STATE \text{\textbf{Add} $P_{notTrain}$ into $SLS$ at random positions}
            \STATE \text{\textbf{Remove} client from $availClients_p$ \textbf{for} $p \in P_{exhaust}$}
        \ENDWHILE

    \STATE $\textcolor{red}{\text{\texttt{\#} Client-side}}$
    \STATE $\theta_{local} \gets \theta_{global} $
    \FOR{$p \in placeholdersToTrain \textbf{ require in } task$}
        \STATE $x,y \gets \text{\textbf{Extract} one $(x,y) \in D_{local}$ where $y=y_{real}$ with}$
        \STATE\hspace{0.2cm} \text{$p \mapsto y_{real}$ if available, else \textbf{add} $p$ to $P_{notTrain}$,\ $P_{exhaust}$}
        \STATE $\theta_{local} \gets \theta_{local} - \eta \cdot \nabla_{\theta_{local}} \mathcal{J}(f_{\theta_{local}}(x), y)$
    \ENDFOR
    \STATE \text{\textbf{Signal} server to send training task to next client in queue}
    \STATE \text{\textbf{Send} updated $\theta$ to next client}
    \STATE \text{\textbf{Return} $P_{notTrain}$, $P_{exhaust}$ back to server}
    
    \end{algorithmic}
\end{algorithm}

\begin{algorithm}[t]
    \small
    \caption{Batch-Data Per Iteration Model Training.}
    \label{alg:blmt}
    \begin{algorithmic}[1] 
        \STATE $\textcolor{red}{\text{\texttt{\#} Server-side}}$
        \STATE $SLS \gets \text{\textbf{Generate} Stratified Label Schedule}$

        \WHILE{$|SLS| \neq 0$}
            \STATE $P_{batch} \gets SLS[0\!:\!batchSize], \ SLS \gets SLS[batchSize\!:]$ 
            \STATE $C_{select} \gets \text{\textbf{Run} Algo. \ref{alg:blcs} on $P_{batch}$}$
            
            \FORALL{$c_i \in \text{set}(C_{select}) \text{, \textbf{concurrently}}$}
                \STATE $P^{c_i}_{subset} \gets [P_{batch}[j] \mid C_{select}[j] = c_i]$
                \STATE $P_{unavail},\ P_{exhaust} \gets \text{\textbf{Send} Training Request containing}$
                \STATE\hspace{3.3cm}$\theta_{global},\ P^{c_i}_{subset}\text{ to }c_i$
                \STATE \text{\textbf{Remove} client from $availClients_p$ \textbf{for} $p \in P_{exhaust}$}
                \STATE \text{\textbf{Rerun} Algo. \ref{alg:blcs} on $P_{unavail}$ and \textbf{Repeat} line 6-9 until}
                \STATE\hspace{0.25cm}\text{available clients are found for all $p \in P_{batch}$}
            \ENDFOR
            \STATE $grad_{sums} \gets \text{\textbf{Signal} awaiting clients to start Local Training}$
            \STATE $grad_{avg} \gets \frac{1}{|P_{batch}|} \sum_{i=0}^{|grad_{sums}|} grad_{sums}[i]$
            \STATE $\theta_{global} \gets \theta_{global} - \eta \cdot grad_{avg}$
        \ENDWHILE

    \STATE $\textcolor{red}{\text{\texttt{\#} Client-side}}$
    \STATE \textbf{Training Request:}
    \STATE $\theta_{local} \gets \theta_{global} $
    \FOR{$p_i \in P^{c_i}_{subset}$}
        \STATE $X_i,Y_i \gets \text{\textbf{Extract} one $(x,y) \in D_{local}$ where $y=y_{real}$ with}$
        \STATE\hspace{0.18cm} \text{$p_i \mapsto y_{real}$ if available, else \textbf{add} $p_i$ to $P_{unavail}$,\ $P_{exhaust}$}
    \ENDFOR
    \STATE \text{\textbf{Return} $P_{unavail}$, $P_{exhaust}$ back to server}
    \STATE \textbf{Local Training:}
    \STATE $\mathcal{J}_{sum} \gets \sum_{i=0}^{|Y|} \ell(f_{\theta_{local}}(X_i), Y_i)$
    \STATE $grad^{local}_{sum} \gets \nabla_{\theta_{local}} \mathcal{J}_{sum}$
    \STATE \text{\textbf{Return} $grad^{local}_{sum}$ to server}
    \end{algorithmic}
\end{algorithm}

Conventional methods (e.g.~FedAvg) rely on aggregating coarse-grained updates after clients train on their entire local datasets for multiple local epochs. Our proposed learning approach presents a novel contribution to the federated learning paradigm by facilitating frequent, fine-grained updates from clients to the global model. This enhances the global model’s ability to adapt quickly and converge more effectively, particularly in heterogeneous settings. 

Following client selection based on the SLS, \shortname\ utilizes two complementary training strategies to update the global model:

(i) \textbf{Single-Sample Learning:} In this learning setting (see Figure \ref{fig_sl} and Algorithm \ref{alg:ssmt}), the global model is passed sequentially from one selected client to the next. At each step, a client performs an update using a single sample corresponding to the label (or consecutive labels) indicated by the SLS. While our approach follows a sequential model-passing structure similar to Sequential Federated Learning (SFL) \cite{li2023convergence}, our method introduces two novel aspects: (1) model updates are routed through a client selection mechanism rather than randomly, and (2) clients perform multiple lightweight updates using small samples each time, rather than full local data training at once. These key aspects significantly enhance the global model's adaptability and performance in non-IID settings. 

(ii) \textbf{Batch-Data Learning:} 
In the batch-data learning setting (see Figure \ref{fig_bl} and Algorithm \ref{alg:blmt}), a mini-batch of samples is aggregated from the selected clients based on the SLS. Each client computes gradients using summed losses over sampled data corresponding to the assigned labels, and these gradients are then aggregated at the server—by summing and normalizing with the batch size—to update the global model. 
This gradient treatment, in conjunction with our guided training and label-aware client selection, ensures that the aggregated gradients remain balanced, unbiased, and exhibit lower variance. Security measures including homomorphic encryption, secure multi-party computation, and differential privacy, can be applied for secure aggregation. We leave the exploration of security aspects as an important direction for future work, since our research primarily focuses on introducing novel guided training approach to address non-IID data.

Our \shortname\ learning approach offers three key benefits:
\begin{itemize}
    \item Fine-Grained, Immediate Updates:
The model receives frequent, incremental updates that immediately incorporate each sample’s contribution, ensuring that no single client’s skewed data disproportionately influences the training.
    \item Mitigation of Non-IID Effects:
By updating the model on a per-sample, label-specific basis, the sequential update mechanism directly counters biases from heterogeneous data distributions. As demonstrated in Section \ref{SLS-A2}, these updates yield an unbiased gradient estimator with reduced variance, leading to improved stability and faster convergence.
Our learning approach is particularly advantageous in scenarios with extreme non-IID data, clearly distinguishing \shortname\ from traditional FL protocols.
    \item Effective participation of clients with limited data: Our learning approach ensures that even a small contribution of clients is meaningfully integrated by accumulating its summed gradients as part of the contribution for a batch or directly incorporates its small update into the single-sample learning model. This addresses a key limitation of conventional FL, which requires clients to possess a sufficient amount of data for effective local training.
\end{itemize}

Together, these two training procedures provide flexibility. Single-sample learning is well-suited in online or streaming scenarios where data arrives sequentially and must be processed within certain time period due to storage constraint. Conversely, batch-data learning is effective when clients have a large and static local data. They can also be used in hybrid, where the global model is first pretrained on large local datasets using batch updates. Once deployed, the model can continuously learn from the new data using single-sample updates, allowing it to adapt over time in dynamic environments.  

\subsection{Custom Batch Normalization Layer}
\label{SLS-A5}
In non-IID FL settings, Batch Normalization (BN) suffers from mismatched batch statistics across clients. This discrepancy can lead to unstable updates and degraded performance \cite{Wang2023}. To cater for the use of BN layer in \shortname, our BN solution is inspired by the solution (FedTAN) proposed in \cite{Wang2023} for FedAvg algorithm. FedTAN modifies the forward and backward propagation of BN layer to compute the global batch statistics and gradients across clients. During the forward pass, the local
batch mean and variance statistics are sent to the server, which averages them to obtain the global batch statistics and send back to clients to proceed with the subsequent computations. Similarly, during the backward pass, an average gradients of these statistics are computed by the server and return back to
clients. 

However, since the forward-pass $\mu$ mean and $\sigma^2$ variance statistics:
\[
    \mu = \frac{1}{m} \sum_{i=1}^{m} x_i, \quad \sigma^2 = \frac{1}{m} \sum_{i=1}^{m} (x_i - \mu)^2,
\]
\noindent and the backward-pass terms in the derivative of the loss $\mathcal{J}$:
\[
    G = \sum_{i=1}^{m} \frac{\partial \mathcal{J}}{\partial \hat{x}_j},
    \quad G_x = \sum_{i=1}^{m} \hat{x}_j \frac{\partial \mathcal{J}}{\partial \hat{x}_j},
\]
\noindent which require batch information involve only sum-based computation. The sums over smaller subsets across clients can be accumulated first, then only divide with the scaling factor. This can better
resolve the mismatch in \shortname, instead of averaging the local batch statistics and gradients across clients to get the global statistics, as the assigned placeholder subsets across clients represent the full batch when combined. Privacy measures such as homomorphic encryption can be applied to shield the batch information from server.

\section{Implementation Details and Communication Efficiency}
\label{Tech-A1}

In practical federated learning systems, securing client participation and managing communication overhead are critical to ensuring system effectiveness. In this section, we describe our privacy-preserving client selection protocol and analyze communication costs, providing real-world implementation details with mathematical rigor.

\subsection{Secure Client Selection with Masked Label Representation}
\label{Tech-A2}

To ensure that only appropriate clients participate in each training iteration while preserving the privacy of their local label distributions. We propose a masked label representation approach that leverages homomorphic encryption (specifically, the CKKS scheme). In this method, each client $i$ computes its local label counts $n_i(l)$ for each label $l$ and then masks these labels and counts by encrypting them as $[E(l_i),E(n_i(l))]$ pairs, using the CKKS encryption scheme. 

To securely identify all unique labels without exposing the actual label value, the server initializes a list of unique labels $\mathcal{L}_{unique}$ by adding all the encrypted labels from first clients.
For each subsequent client $i$, the server compares its encrypted label $E(l_i)$ with the labels already in $\mathcal{L}_{unique}$ using encrypted subtraction operation:
\[ E(\text{Diff}_{ij}) =  E(l_i) - E(l_j), \ \forall E(l_j) \in \mathcal{L}_{\text{unique}}.\]
\noindent The result of this operation against all unique labels is decrypted by client $i$. If they are all not 0, $E(l_i)$ is considered unique and added to $\mathcal{L}_{unique}$. Once all unique labels are identified, each unique $E(l)$ is mapped to a randomly generated placeholder $p$ (e.g., ASCII codes) to facilitate the subsequent clients label-to-placeholder mapping. The mapping process uses also the encrypted subtraction operation by comparing each client encrypted label $E(l_i)$ with the unique labels in $\mathcal{L}_{unique}$. If a match is found, the corresponding placeholder $p$ is assigned to that client's label. 

With all labels mapped to placeholders, the server proceeds to aggregate the encrypted counts $E(n_i(l))$ from all clients to obtain the global count for each placeholder:
\[
E(N(p)) = \sum_{i \in \mathcal{C}_p} E(n_i(l)),\ E(l_i) \mapsto p,
\]
\noindent where $\mathcal{C}_p$ denotes the set of clients with $E(l_i)$ mapped to $p$. Thanks to the additive homomorphic properties of CKKS, the server can compute $E(N(p))$ without decrypting individual values. Once aggregation is complete, the server, in coordination with the clients (who hold the decryption keys), securely decrypts $E(N(p))$ to reveal the global counts $N(p)$ in a privacy-preserving manner. These counts then inform the frequency $f_l$ used in constructing the SLS (see Section \ref{SLS-A1}).

This unified protocol combining masked label representation with homomorphic encryption ensures accurate global label statistics while protecting client privacy.

\subsection{Communication Overhead Analysis}
\label{Tech-A3}

Efficient communication is essential for practical federated learning. Let $T$ denotes the total communication time per client, $R$ denotes the number of training rounds, $C_{\text{update}}$ represents the cost (in seconds or data volume) per model update (weights or gradients) transfer, and $f_{\text{freq}}$ represent the frequency of model update transfers per round:
\[ T = R \times C_{\text{update}} \times f_{\text{freq}}.\]

Since \shortname\ employs frequent, fine-grained global model updates, $f_{\text{freq}}$ is higher in our learning approach compared to conventional methods. Specifically, in our batch-learning approach, the transfer frequency is once per batch iteration, assuming that the client participates in every batch iterations. Whereas in our single-sample approach, the transfer frequency per client is equivalent to the expected amount of local labels contributing to the SLS, when the $chunkSize$ is set to 1. As the $chunkSize$ increases, $f_{\text{freq}}$ reduces due to consecutive placeholders being assigned to the same client per training task (see ablation study in Section \ref{ablationStudy}). 

Although $f_{\text{freq}}$ is higher in our learning approach, increasing the number of clients effectively spreads out $f_{\text{freq}}$ to other available clients, leading to lower $f_{\text{freq}}$ per client. Empirical findings further indicate that our learning approach requires significantly fewer communication rounds ($R$) to reach convergence compared to all baselines. In the federated learning scenarios where high-bandwidth connections are available, the additional communication overhead is often negligible. 

\subsection{Integration with \shortname\ Training Procedures}
\label{Tech-A4}

The practical implementation details described above ensure that the theoretical benefits of our guided training mechanism are preserved during real-world operation. Specifically, the secure client selection protocol with its masked label representation enabled by CKKS homomorphic encryption guarantees that the SLS is constructed accurately without compromising client privacy. The communication overhead analysis demonstrates that, despite the higher update frequency inherent in \shortname\ learning strategy, the overall communication cost is kept manageable due to the reduced number of training rounds.

These implementation aspects bridge the gap between our theoretical framework (Sections \ref{SLS-A1}-\ref{SLS-A2}) and practical deployment, confirming that \shortname\ is both effective in theory and viable in real-world federated learning scenarios.

\begin{table*}[!t]
    \caption{Averaged Top-1 Test Accuracy of Federated Learning Algorithms using Single-Sample Per Iteration Learning (Batch Size 1) with 10 Clients. E\# denotes the epoch at which the algorithm achieves its highest accuracy. U and W denote the accuracy is achieved using uniform or weighted client selection. \label{tab:ss_result}}.
    \centering
    \footnotesize
    \begin{tabular}{ccccccccc}
        \toprule
        \textbf{Non-IID} & \textbf{Dataset} & \textbf{Partitioning} & \textbf{FedAvg} & \textbf{FedProx} & \textbf{SCAFFOLD} & \textbf{SFL} & \textbf{\shortname\ (Ours)} \\
        \midrule
        
        \multirow{2}{*}{Label Skew} & \multirow{2}{*}{MNIST} 
        & \#C=1 & 18.71\%, E10 & 12.37\%, E14 & 17.68\%, E30 & 15.19\%, E29 & \textbf{98.80\%, E4} \\
        & & \#C=3 & 91.67\%, E30 & 85.27\%, E29 & 91.11\%, E18 & 72.52\%, E30 & \textbf{99.06\%, E5} \\
        & & \#C=5 & 95.42\%, E30 & 96.51\%, E30 & 92.60\%, E24 & 89.30\%, E28 & \textbf{98.95\%, E5} \\
        & & Dirichlet(0.5) & 97.82\%, E30 & 97.19\%, E30 & 98.75\%, E27 & 97.76\%, E28 & \textbf{98.94\%, E5} \\
        \cmidrule(lr){3-8}
         & & IID & 98.32\%, E30 & 98.34\%, E30 & 98.71\%, E19 & 98.79\%, E5 & 98.82\%, E5 \\
        \cmidrule{2-8}
        
        & \multirow{2}{*}{CIFAR-10} 
        & \#C=1 & 11.56\%, E7 & 11.43\%, E38 & 10.46\%, E2 & 10.00\%, E1 & \textbf{81.18\%, E29} \\
        & & \#C=3 & 42.61\%, E50 & 50.44\%, E49 & 59.12\%, E50 & 30.10\%, E42 & \textbf{80.15\%, E22} \\
        & & \#C=5 & 66.36\%, E50 & 65.94\%, E50 & 73.83\%, E50 & 60.10\%, E46 & \textbf{80.71\%, E30} \\
        & & Dirichlet(0.5) & 72.37\%, E50 & 66.75\%, E50 & 75.13\%, E50 & 76.38\%, E24 & \textbf{81.06\%, E23} \\
        \cmidrule(lr){3-8}
         & & IID & 73.76\%, E50 & 74.28\%, E50 & 74.48\%, E50 & 82.60\%, E50 & 80.15\%, E30 \\
        \cmidrule{2-8}
        
        & \multirow{2}{*}{CIFAR-100} 
        & \#C=10 & 32.55\%, E50 & 18.77\%, E49 & 27.74\%, E50 & 12.53\%, E46 & \textbf{55.00\%, E43} \\
        & & \#C=30 & 35.33\%, E50 & 34.92\%, E50 & 32.52\%, E49 & 31.22\%, E48 & \textbf{54.32\%, E41} \\
        & & \#C=50 & 37.12\%, E49 & 36.57\%, E50 & 33.13\%, E50 & 44.88\%, E48 & \textbf{54.69\%, E48} \\
        & & Dirichlet(0.5) & 38.56\%, E50 & 38.50\%, E50 & 33.76\%, E50 & 51.80\%, E45 & \textbf{54.62\%, E45} \\
        \cmidrule(lr){3-8}
         & & IID & 39.6\%, E50 & 39.8\%, E50 & 35.71\%, E50 & 63.22\%, E50 & 54.26\%, E50 \\
        \cmidrule{2-8}
        
        & \multirow{2}{*}{COVTYPE} 
        &  \#C=1 & 58.61\%, E22 & 54.34\%, E30 & 57.07\%, E1 & 57.03\%, E1 & \textbf{87.07\%, E30} \\
        & & Dirichlet(0.5) & 75.07\%, E27 & 66.24\%, E28 & 71.60\%, E24 & 44.10\%, E12 & \textbf{86.19\%, E28} \\
        \cmidrule(lr){3-8}
         & & IID & 77.38\%, E30 & 70.64\%, E30 & 60.00\%, E30 & 86.65\%, E30 & 86.94\%, E30 \\
         
        \midrule
        \multirow{2}{*}{Feature Skew} & \multirow{2}{*}{PACS} 
        &  \#D=1 & 87.13\%, E29 & 92.00\%, E30 & 88.60\%, E30 & 94.93\%, E30 & \textbf{95.24\%, E15(U)} \\
        & & \#D=2 & 94.65\%, E28 & 95.28\%, E30 & 89.31\%, E30 & \textbf{95.61\%, E29} & 95.35\%, E13(U) \\
        & & Dirichlet(0.5) & 84.4\%, E30 & 90.74\%, E29 & 85.68\%, E30 & 89.57\%, E29 & \textbf{95.84\%, E14(U)} \\
        \cmidrule(lr){3-8}
         & & IID & 95.72\%, E30 & 95.68\%, E30 & 92.57\%, E30 & 95.30\%, E30 & 95.12\%, E15 \\
        \cmidrule{2-8}
        
        & \multirow{2}{*}{Digits-DG} 
        &  \#D=1 & 82.69\%, E15 & 82.56\%, E15 & \textbf{83.04\%, E13} & 78.60\%, E13 & 82.48\%, E13(U) \\
        & & \#D=2 & 83.09\%, E13 & \textbf{83.95\%, E15} & 82.83\%, E14 & 82.14\%, E14 & 82.33\%, E14(U) \\
        & & Dirichlet(0.5) & 82.31\%, E15 & \textbf{83.56\%, E15} & 82.83\%, E15 & 78.00\%, E7 & 82.6\%, E15(U) \\
        \cmidrule(lr){3-8}
        & & IID & 84.44\%, E15 & 83.79\%, E15 & 83.56\%, E15 & 81.79\%, E15 & 82.20\%, E15 \\
        
        \bottomrule
    \end{tabular}
\end{table*}

\begin{table*}[!t]
    \caption{Averaged Top-1 Test Accuracy of Federated Learning Algorithms using Batch-Data Per Iteration Learning with 10 Clients. \\E\# denotes the epoch at which the algorithm achieves its highest accuracy. U and W denote the accuracy is achieved using uniform or weighted client selection.\label{tab:bl_result}}
    \centering
    \footnotesize
    \begin{tabular}{ccccccccc}
        \toprule
        \textbf{Non-IID} & \textbf{Dataset} & \textbf{Partitioning} & \textbf{FedAvg} & \textbf{FedProx} & \textbf{SCAFFOLD} & \textbf{\shortname\ (Ours)} \\
        \midrule
        
        \multirow{2}{*}{Label Skew} & \multirow{2}{*}{\shortstack{CIFAR-10 \\ (with BN)}} 
        & \#C=1 & 23.45\%, E44 & 29.38\%, E49 & 18.27\%, E36 & \textbf{90.60\%, E8}  \\
        & & \#C=3 & 62.15\%, E39 & 52.75\%, E50 & 66.72\%, E50 & \textbf{90.45\%, E8} \\
        & & \#C=5 & 64.03\%, E45 & 71.40\%, E50 & 84.39\%, E47 & \textbf{90.54\%, E8} \\
        & & Dirichlet(0.5) & 81.29\%, E42 & 80.11\%, E47 & 84.54\%, E48 & \textbf{90.27\%, E8} \\
        \cmidrule(lr){3-7}
         & & IID & 88.84\%, E50 & 88.71\%, E50 & 87.48\%, E50 & 90.39\%, E8 \\
        \cmidrule{2-7}
        
        & \multirow{2}{*}{\shortstack{CIFAR-100 \\ (with BN)}} 
        & \#C=10 & 36.40\%, E50 & 36.73\%, E48 & 39.10\%, E42 & \textbf{74.45\%, E30} \\
        & & \#C=30 & 42.39\%, E50 & 42.18\%, E50 & 51.76\%, E48 & \textbf{74.80\%, E29} \\
        & & \#C=50 & 46.77\%, E50 & 46.80\%, E50 & 61.39\%, E49 & \textbf{74.65\%, E30} \\
        & & Dirichlet(0.5) & 47.57\%, E50 & 46.62\%, E50 & 64.74\%, E49 & \textbf{74.72\%, E30} \\
        \cmidrule(lr){3-7}
         & & IID & 48.28\%, E50 & 48.64\%, E50 & 66.19\%, E50 & 74.41\%, E30 \\
        \cmidrule{2-7}
        
        & \multirow{2}{*}{\shortstack{Tiny-ImageNet \\ (with BN)}} 
        & \#C=20 & 14.93\%, E70 & 12.65\%, E70 & 25.88\%, E70 & \textbf{51.09\%, E50}  \\
        & & \#C=60 & 17.86\%, E70 & 17.85\%, E70 & 35.10\%, E70 & \textbf{51.60\%, E49}  \\
        & & \#C=100 & 22.14\%, E70 & 21.36\%, E70 & 41.28\%, E70 & \textbf{50.72\%, E49}  \\
        & & Dirichlet(0.5) & 23.10\%, E70 & 23.21\%, E70 & 42.40\%, E68 & \textbf{50.62\%, E46}  \\
        \cmidrule(lr){3-7}
         & & IID & 35.69\%, E70 & 35.59\%, E70 & 43.77\%, E70 & 51.15\%, E50  \\
        \cmidrule{2-7}
        
        & \multirow{2}{*}{COVTYPE} 
        &  \#C=1 & 57.03\%, E1 & 57.01\%, E1 & 57.31\%, E25 & \textbf{90.19\%, E29}  \\
        & & Dirichlet(0.5) & 77.26\%, E39 & 80.23\%, E49 & 78.14\%, E46 & \textbf{89.95\%, E27}  \\
        \cmidrule(lr){3-7}
         & & IID & 85.60\%, E50 & 85.75\%, E50 & 84.17\%, E50 & 89.48\%, E30 \\
         
        \midrule
        \multirow{2}{*}{Feature Skew} & \multirow{2}{*}{PACS} 
        &  \#D=1 & 78.69\%, E30 & 85.85\%, E29 & 42.67\%, E13 & \textbf{94.74\%, E7(W)}  \\
        & & \#D=2 & 93.26\%, E30 & 82.57\%, E30 & 92.32\%, E30 & \textbf{94.24\%, E7(W)} \\
        & & Dirichlet(0.5) & 76.04\%, E30 & 74.59\%, E30 & 77.59\%, E30 & \textbf{95.71\%, E7(W)} \\
        \cmidrule(lr){3-7}
         & & IID & 93.80\%, E30 & 93.92\%, E30 & 90.88\%, E30 & 94.15\%, E7 \\
        \cmidrule{2-7}
        
        & \multirow{2}{*}{Digits-DG} 
        &  \#D=1 & 78.31\%, E15 & 80.48\%, E15 & 68.94\%, E15 & \textbf{84.75\%, E13(U)} \\
        & & \#D=2 & 80.92\%, E15 & 79.77\%, E15 & 75.39\%, E15 & \textbf{85.03\%, E11(U)} \\
        & & Dirichlet(0.5) & 77.96\%, E14 & 78.46\%, E15 & 56.02\%, E15 & \textbf{84.73\%, E13(U)} \\
        \cmidrule(lr){3-7}
        & & IID & 80.21\%, E15 & 81.60\%, E15 & 74.73\%, E15 & 85.21\%, E13 \\
        
        \bottomrule
    \end{tabular}
\end{table*}

\section{Experimental Results and Discussion}

In this section, we present extensive experimental evaluations of our proposed \shortname\ framework on multiple benchmark datasets. We compare \shortname\ with several state‑of‑the‑art methods under diverse non‑IID conditions and analyze key aspects such as convergence behavior, accuracy, and communication overhead.

 Baseline FL algorithms, including FedAvg, FedProx, SCAFFOLD, vanilla SFL, and FedTAN are implemented for comparisons. The global model performance is evaluated using the average Top-1 test accuracy across all clients. The algorithms are trained for a maximum epoch took in their respective IID case, and the epoch that obtains the highest accuracy is recorded. We evaluated \shortname\ on datasets including MNIST, CIFAR‑10, CIFAR‑100, Tiny‑ImageNet, and COVTYPE (for label skew) as well as PACS and Digits‑DG (for feature skew). This study follows the non-IID data partitioning methods proposed in \cite{li2022federated}, where datasets are partitioned based on the number of classes (\#C) or domains (\#D) assigned to each client. The severity of non-IID increases as the number of assigned classes or domains decreases. The samples of each label are randomly and equally distributed to the clients that hold the label. To simulate imbalanced data distributions, a common challenge in federated learning, we partitioned the datasets using both label‑skew schemes and a Dirichlet distribution (with a concentration parameter of 0.5). The Dirichlet distribution is widely used to control the degree of heterogeneity among clients; a lower concentration parameter (0.5) results in highly skewed, non‑IID data. This isolates the impact of client heterogeneity without confounding effects from global class imbalance. 

For single-sample learning, we utilized simple CNN models across all datasets. For the batch-data learning, we employed ResNet-9 with BN for CIFAR-10, CIFAR-100, and ResNet-18 with BN for Tiny-ImageNet datasets; and utilized simple CNN models for COVTYPE, PACS, and Digits-DG. \shortname\ employs uniform client selection across all non-IID scenarios for the label skew setting. In the feature skew setting, both uniform and weighted client selection strategies are explored, with the better-performing one reported.

\subsection{Impact of Non-IID Severity on Performance}
As shown in Table \ref{tab:ss_result} and \ref{tab:bl_result}, as the number of labels (\#C) and domains (\#D) each client holds decreases, \shortname\ is the only algorithm that consistently achieves accuracy close to IID case in both single-sample and batch-data learning settings, while the accuracy of all baselines decline in most cases. In Dirichlet-based data partition scenarios, \shortname\ consistently outperforms all baselines, particularly in batch-data learning. Its performance remain stable even in the presence of label imbalance. \shortname's 
 resiliency towards varying non-IID severity is attributed to the use of $SLS$ that restricts the global model to be updated following the label sequence defined. It iteratively takes a fine-grained and immediate global update strategy using $SLS$ as its guide, which enables the global model to adapt to the diverse data in a gradual and unbiased manner. 

On the contrary, all baselines require clients to train on most or all of their available data at once before sending the update to server. This results in diverged local updates in non-IID settings as each client's update is constructed based on a skewed data distribution. While FedProx and SCAFFOLD introduce mechanisms to mitigate local drift, their correction effectiveness is limited in highly non-IID settings and is inconsistent across the two learning settings and non-IID types. SCAFFOLD, while more effective than FedProx on label-skew data with batch learning setting, struggles in single-sample learning and perform poorly with feature-skew data. This is because SCAFFOLD's control variates can only adjust the gradient drift in clients' updates to ensure update consistency, which does not explicitly align drifted feature representations. 

\shortname\ effectively addresses feature skew by consistently introducing the global model to label data from diverse clients using the client selection mechanisms that designed to ensure an unbiased feature representations learning. FedProx, on the other hand, requires a proper tuning of proximal coefficient based on the degree of data heterogeneity, which is often unknown in practice. As a result, finding a suitable coefficient requires multiple rounds of FL, making the optimization process less efficient with unguaranteed performance improvement on severe data heterogeneity scenarios. These show that correcting local drift is challenging, as its effectiveness depends on the degree of data heterogeneity and a proper parameter tuning. Hence, without addressing the fundamental design limitations of FedAvg in handling non-IID data, these baselines struggle to consistently improve performance across different non-IID scenarios. 

\subsection{Performance with Increasing Dataset Classes}
Among all the evaluated algorithms, \shortname\ is the only algorithm that consistently achieves accuracy similar to its IID case across all datasets with varying class sizes. This demonstrates \shortname's robustness in learning from a broader range of classes by incorporating all the diverse labels in $SLS$. In contrast, the accuracy gap between FedAvg, FedProx, and SCAFFOLD compared to \shortname\ increases in both single-sample and batch-data label-skew settings as the number of dataset classes increases. This trend can be clearly observed by analyzing the accuracies across datasets partitioned using Dirichlet(0.5) and the least severe \#C, both of which exhibit less extreme label-skew to better isolate the effect of increasing dataset classes. SFL, on the other hand, manages to maintain a small accuracy gap in the Dirichlet(0.5) partitioned due to its sequential learning nature, which allows it to progressively learn across variable range of labels. However, this would require SFL clients to have minimal non-overlapping labels to minimize catastrophic forgetting. In the \#C partition type, where clients have larger disjoint label sets, the accuracy gap between SFL and \shortname\ widens as the learned representations from previous clients are progressively overwritten by recent clients with disjoint labels. 

Hence, \shortname\ proves to be a more effective approach for scenarios with increasing class diversity, as it consistently maintain high accuracy where the baselines struggle. 

\subsection{Performance with Increasing Client Size}
\begin{figure*}[!t]
    \centering
    \hspace{-18pt}
    \subfloat[Single-sample, CIFAR-10, \#C=3]{\includegraphics[width=6.0cm, height=4.5cm]{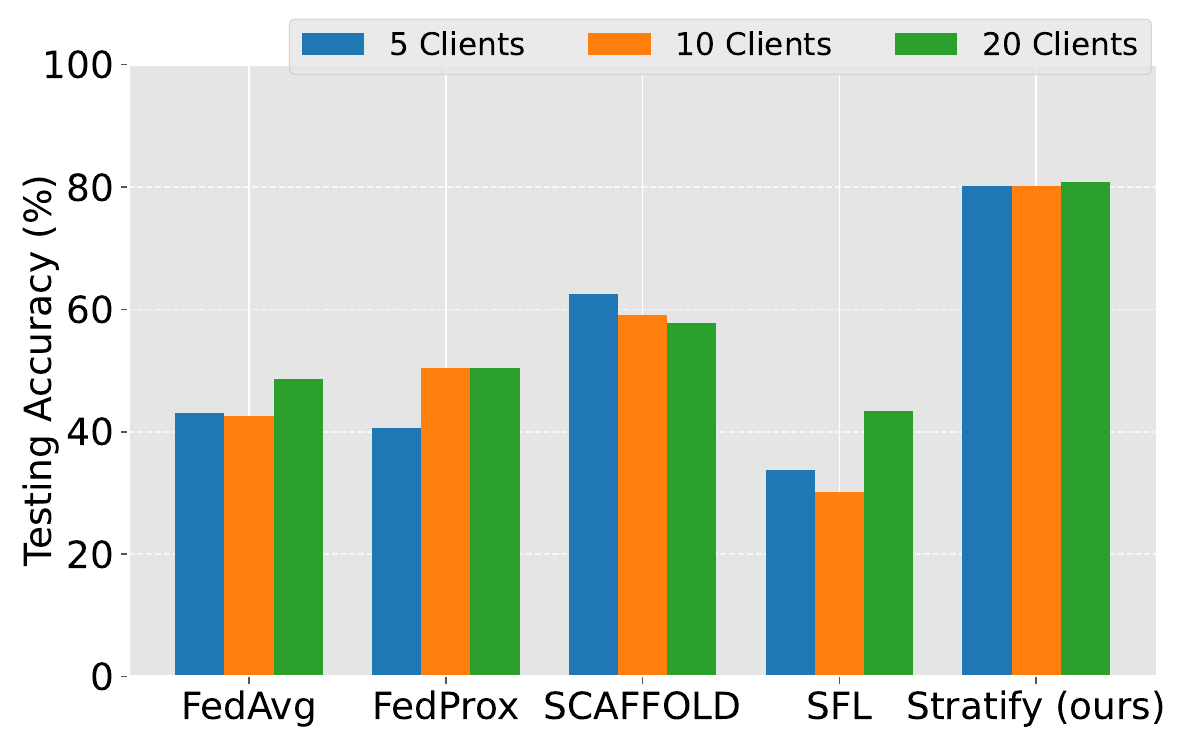}%
    \label{fig:chart1}}
    \hspace{-2pt}
    \subfloat[Single-sample, CIFAR-100, \#C=50]{\includegraphics[width=6.0cm, height=4.5cm]{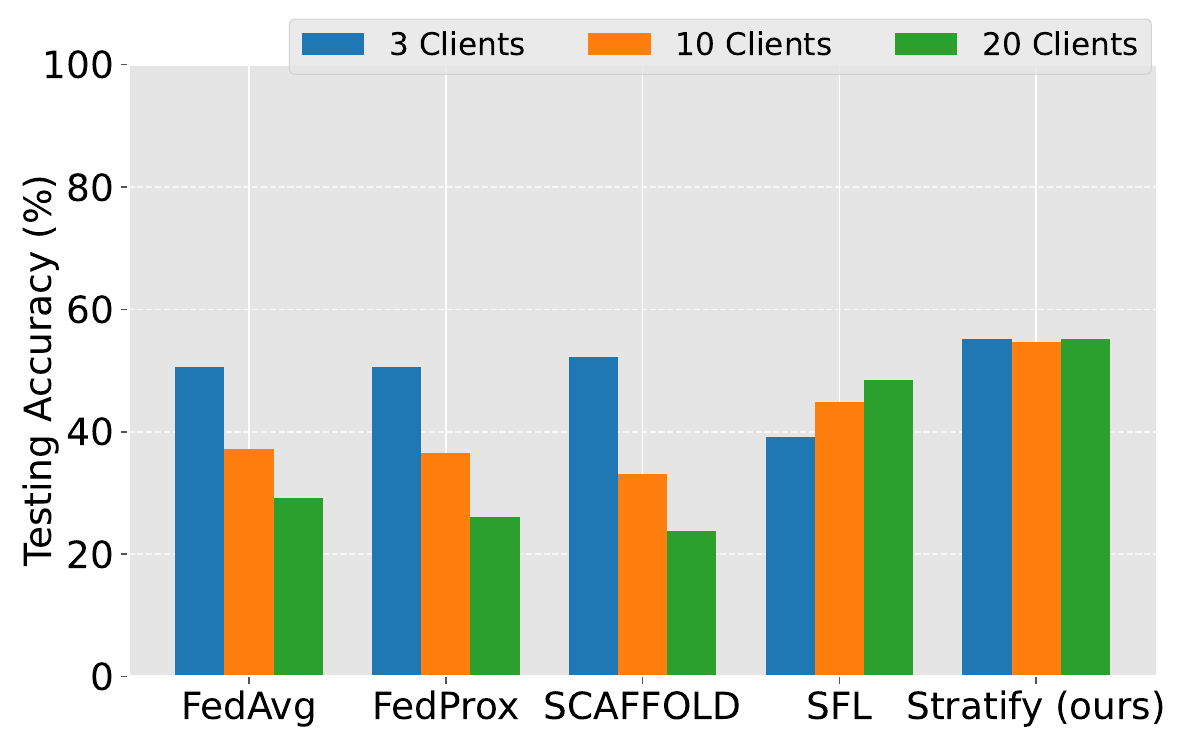}%
    \label{fig:chart2}}
    \hspace{-2pt}
    \subfloat[Single-sample, PACS, \#D=1]{\includegraphics[width=6.0cm, height=4.5cm]{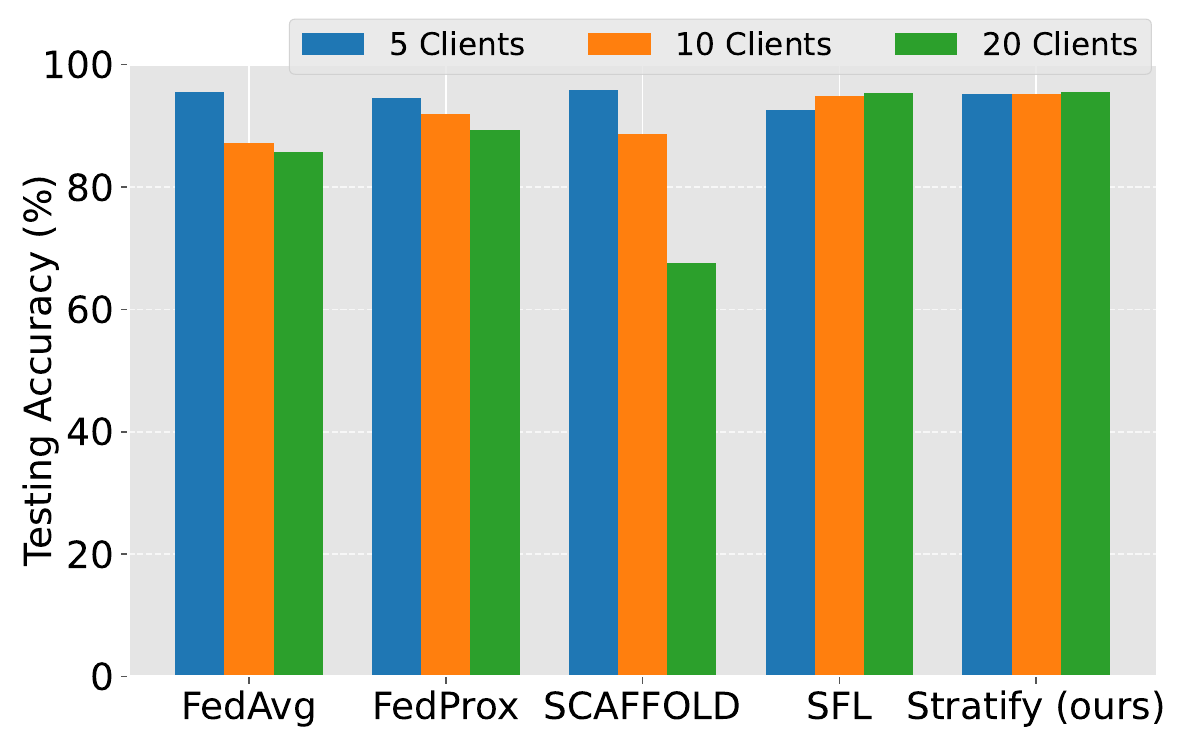}
    \label{fig:chart3}}
    \hspace{-10pt}

    \vspace{10pt}
    \vfill
    \hspace{-18pt}
    \subfloat[Single-sample, CIFAR-10, Dir(0.5)]{\includegraphics[width=6.0cm, height=4.5cm]{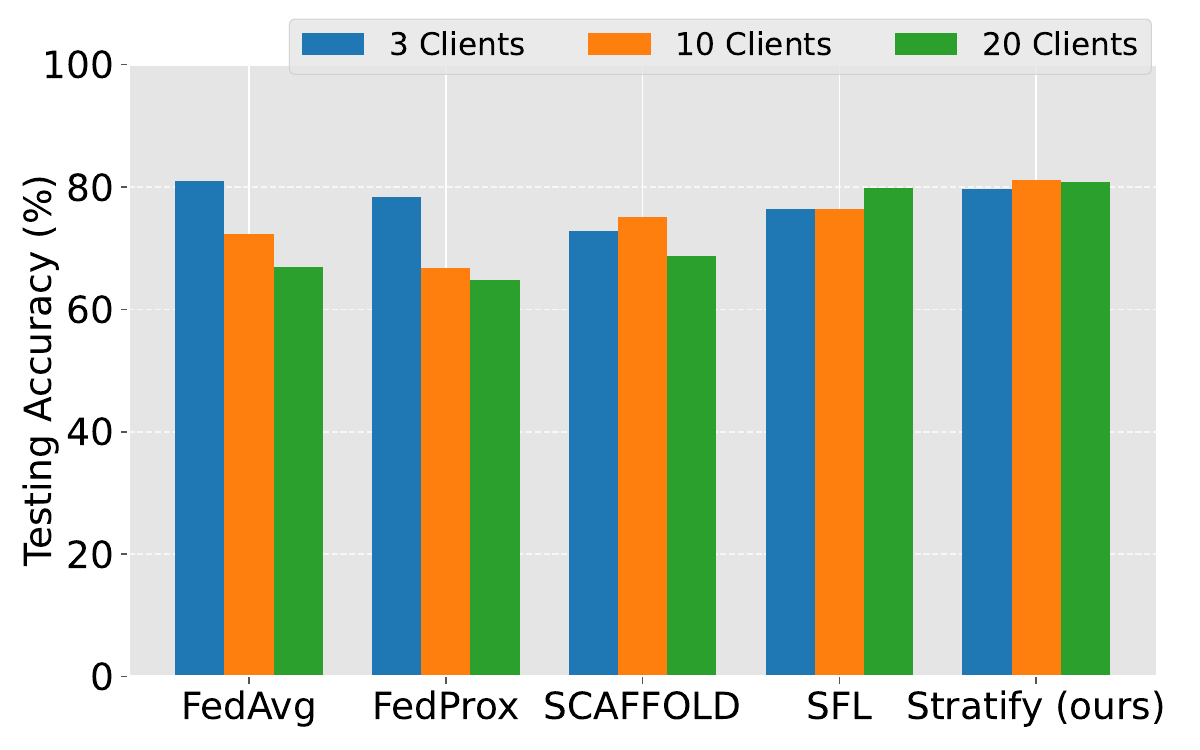}%
    \label{fig:chart4}}
    \hspace{-2pt}
    \subfloat[Single-sample, CIFAR-100, Dir(0.5)]{\includegraphics[width=6.0cm, height=4.5cm]{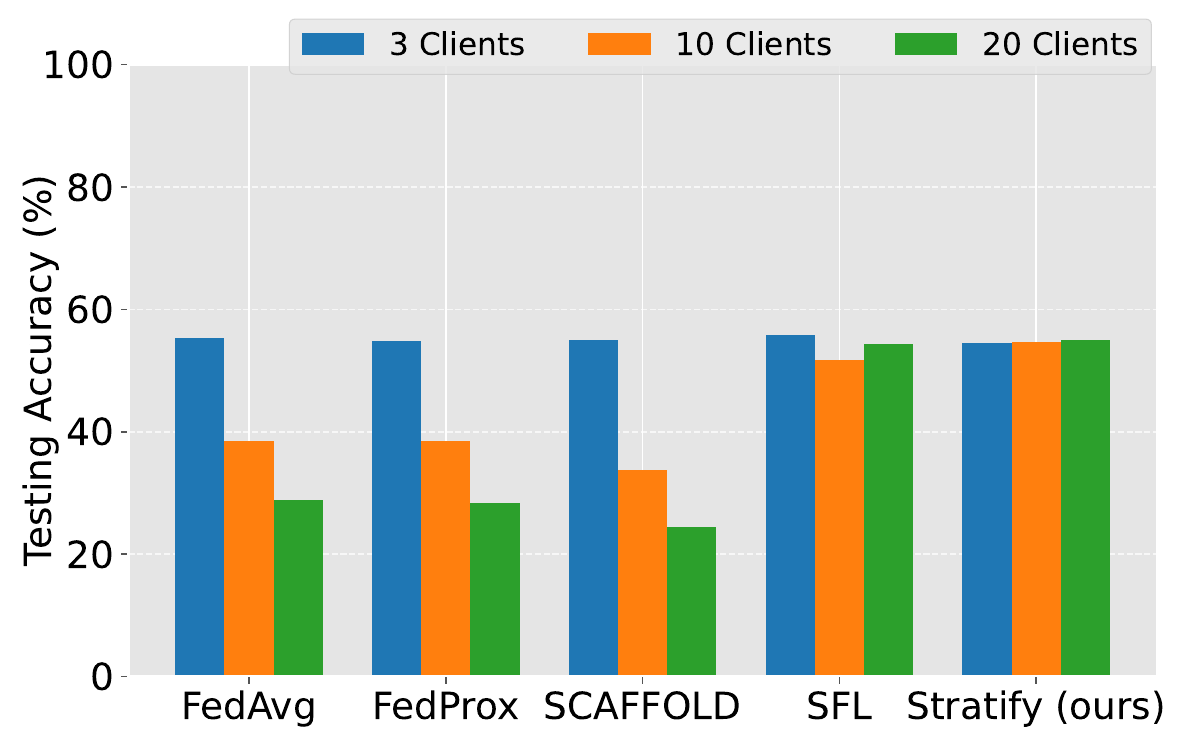}%
    \label{fig:chart5}}
    \hspace{-2pt}
    \subfloat[Single-sample, PACS, Dir(0.5)]{\includegraphics[width=6.0cm, height=4.5cm]{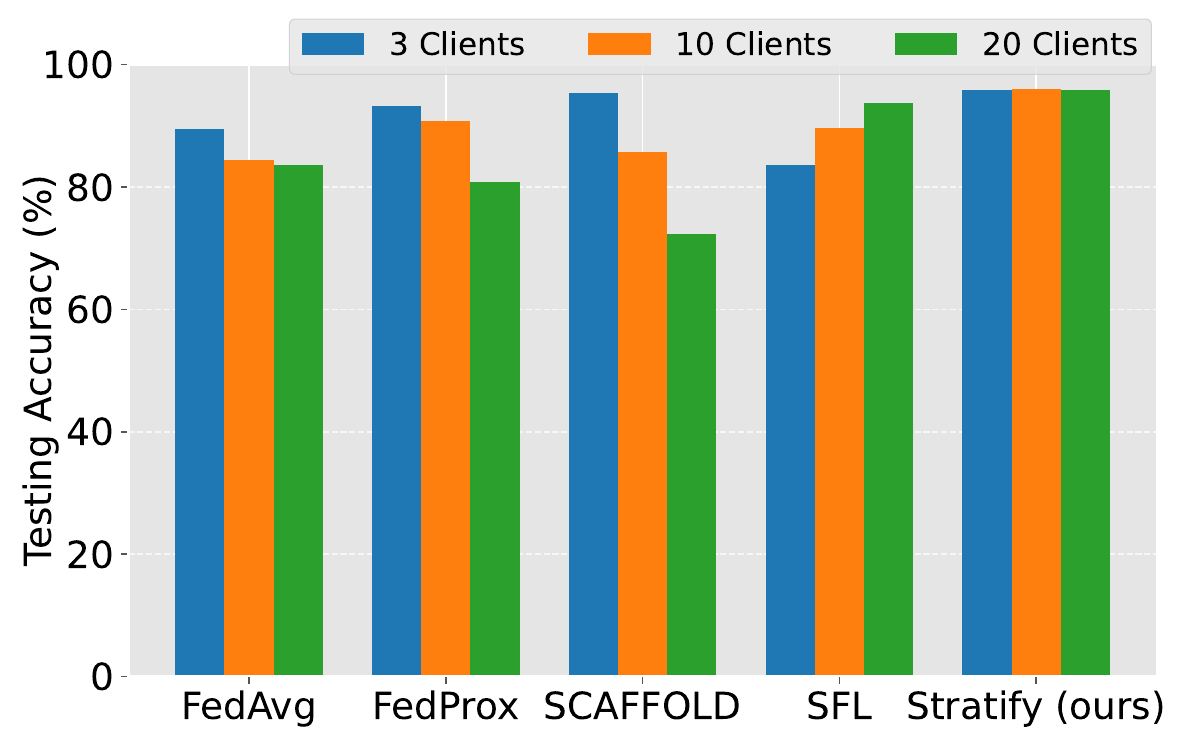}%
    \label{fig:chart6}}
    \hspace{-10pt}
    
    \caption{Comparison of algorithms' performance with increasing client numbers on different datasets and data partitions in single-sample learning}
    \label{fig:bar_charts}
\end{figure*}

\begin{figure*}[!t]
    \centering
    \vspace{10pt}
    \vfill
    \hspace{-18pt}
    \subfloat[Batch-data, CIFAR-10, \#C=3]{\includegraphics[width=6.0cm, height=4.5cm]{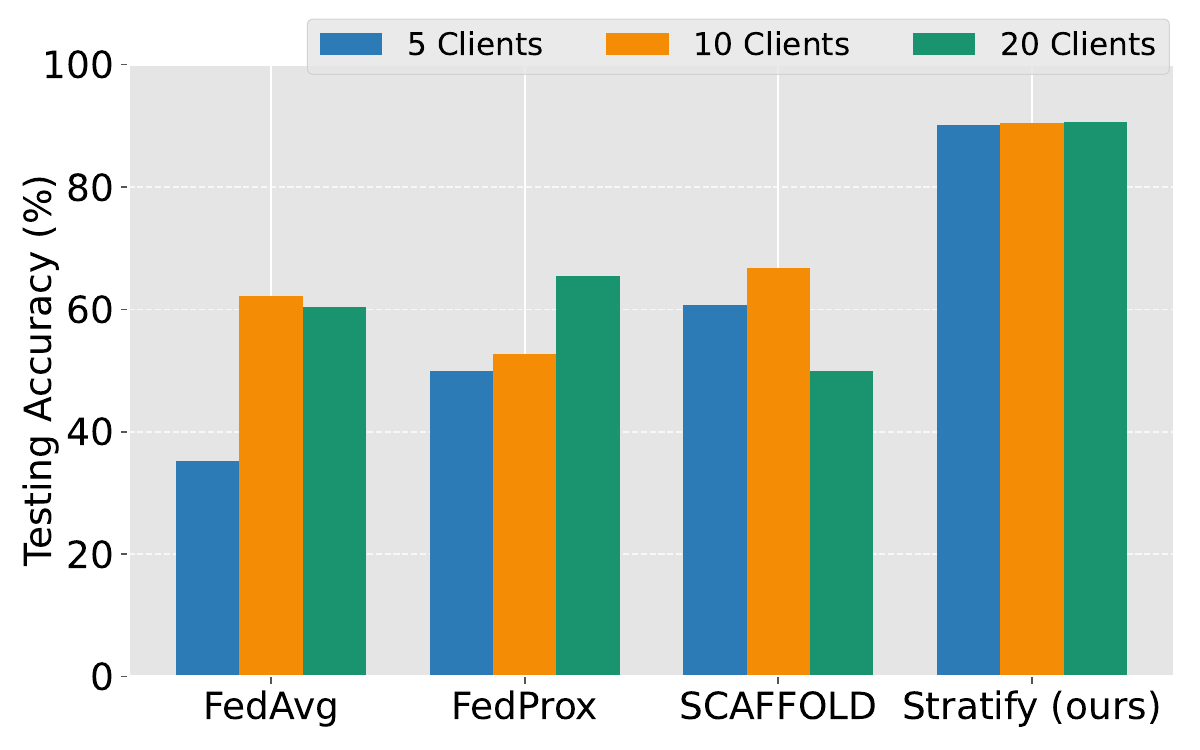}%
    \label{fig:chart7}}
    \hspace{-2pt}
    \subfloat[Batch-data,  CIFAR-100, \#C=50]{\includegraphics[width=6.0cm, height=4.5cm]{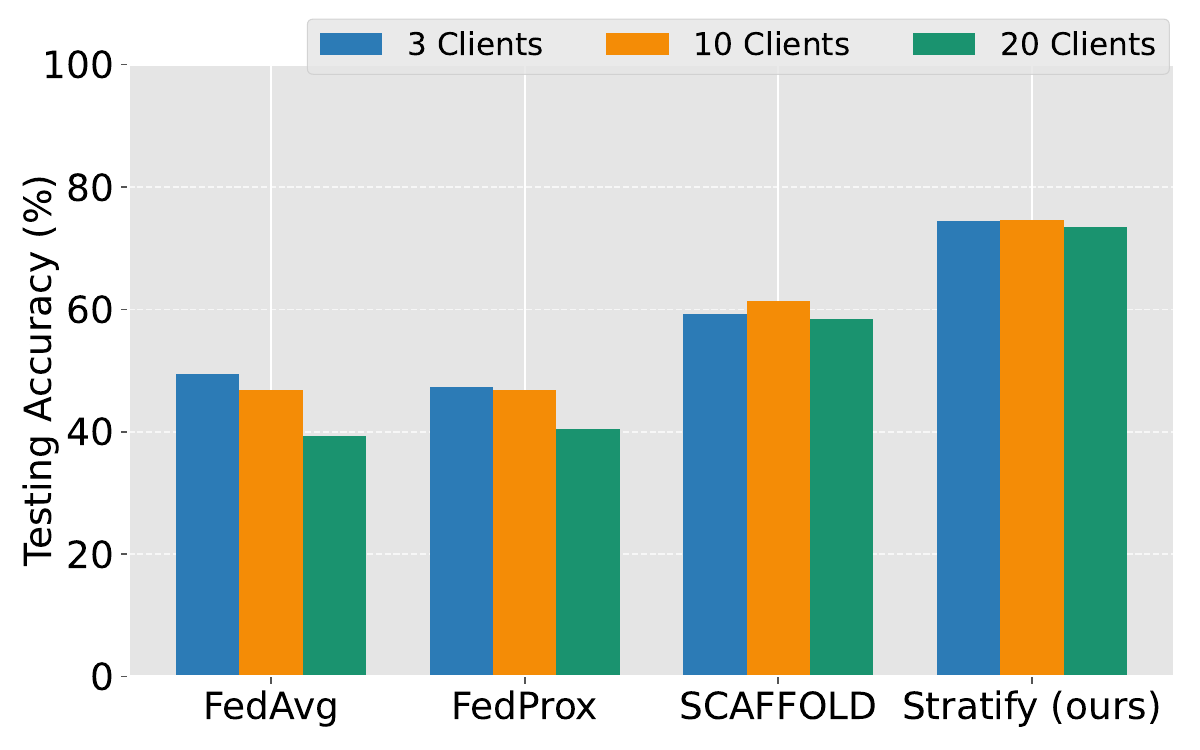}%
    \label{fig:chart8}}
    \hspace{-2pt}
    \subfloat[Batch-data, PACS, \#D=1]{\includegraphics[width=6.0cm, height=4.5cm]{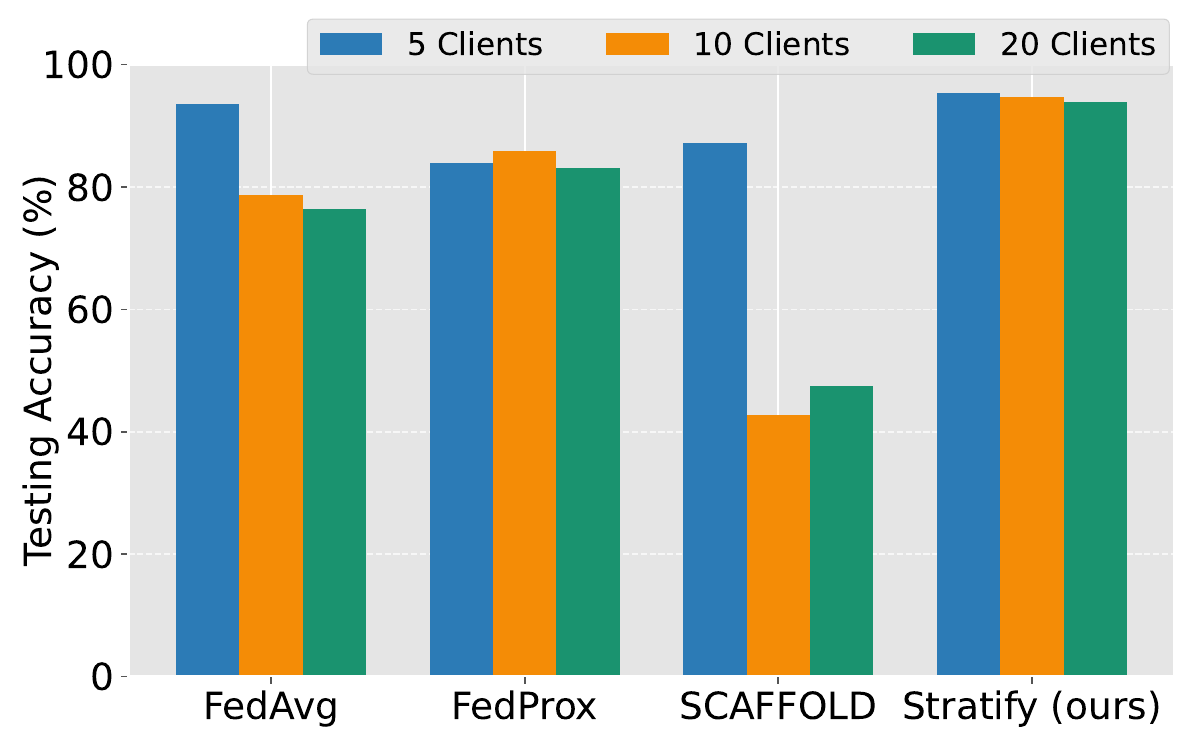}%
    \label{fig:chart9}}
    \hspace{-10pt}

    \vspace{10pt}
    \vfill
    \hspace{-18pt}
    \subfloat[Batch-data, CIFAR-10, Dir(0.5)]{\includegraphics[width=6.0cm, height=4.5cm]{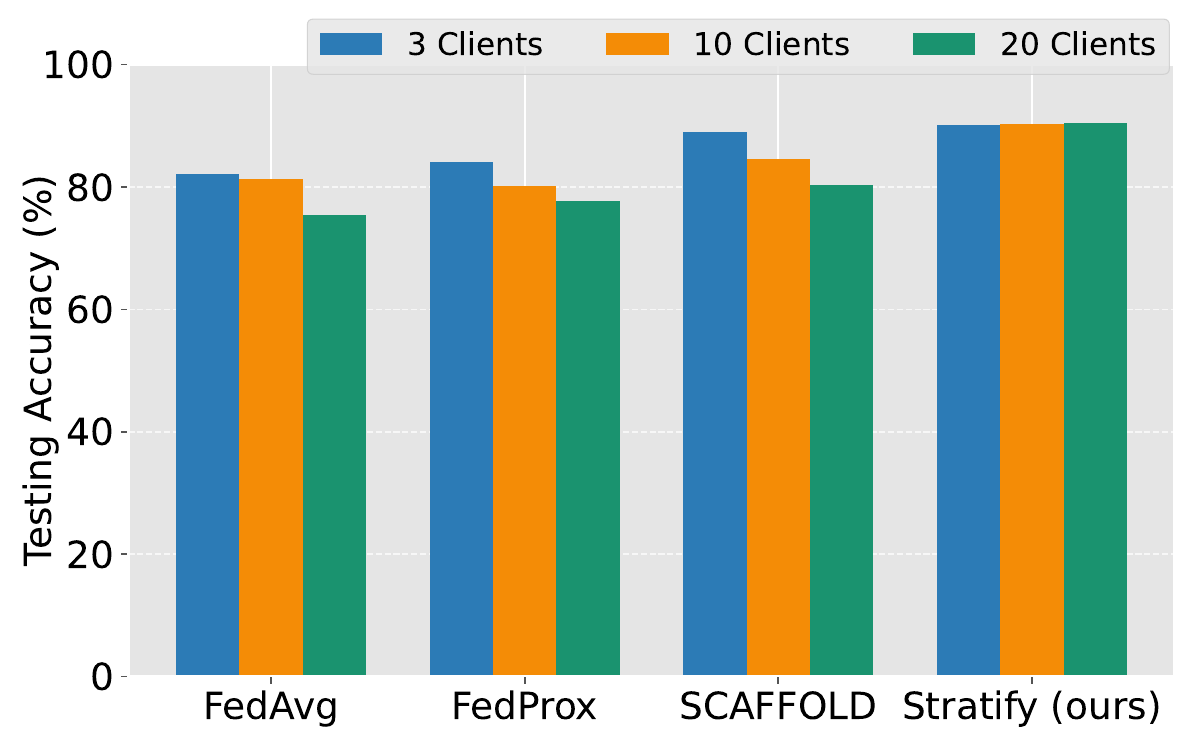}%
    \label{fig:chart10}}
    \hspace{-2pt}
    \subfloat[Batch-data, CIFAR-100, Dir(0.5)]{\includegraphics[width=6.0cm, height=4.5cm]{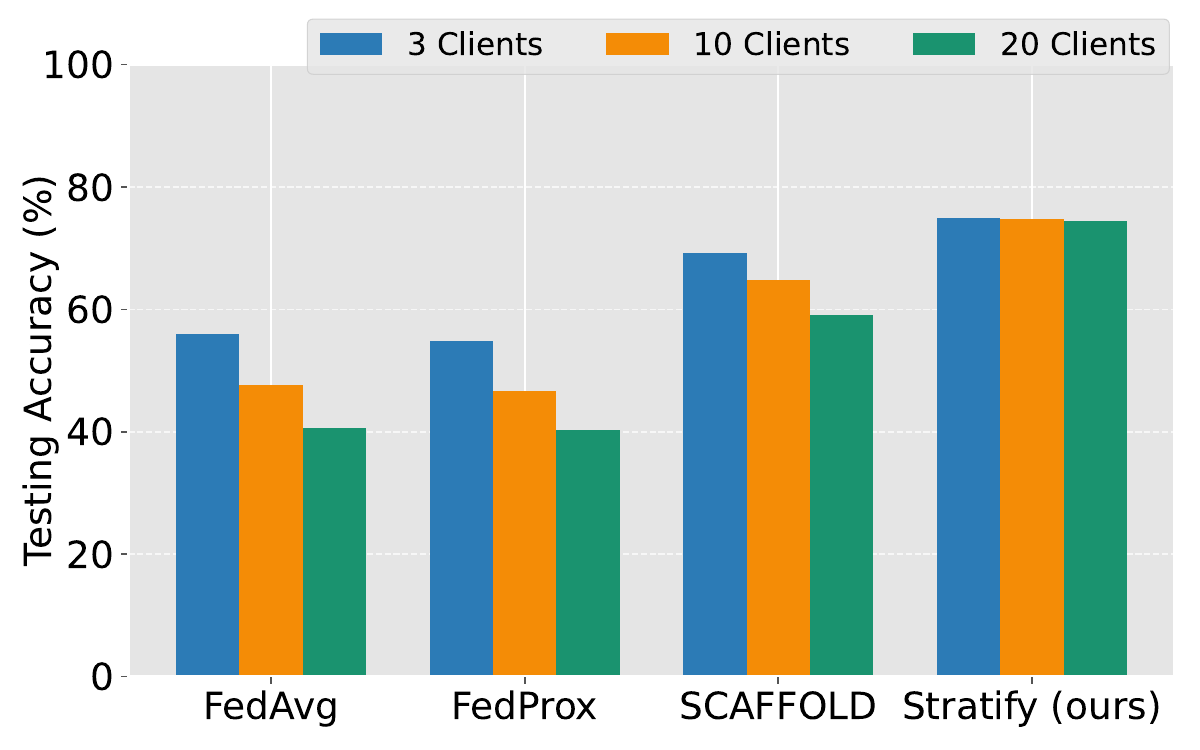}%
    \label{fig:chart11}}
    \hspace{-2pt}
    \subfloat[Batch-data, PACS, Dir(0.5)]{\includegraphics[width=6.0cm, height=4.5cm]{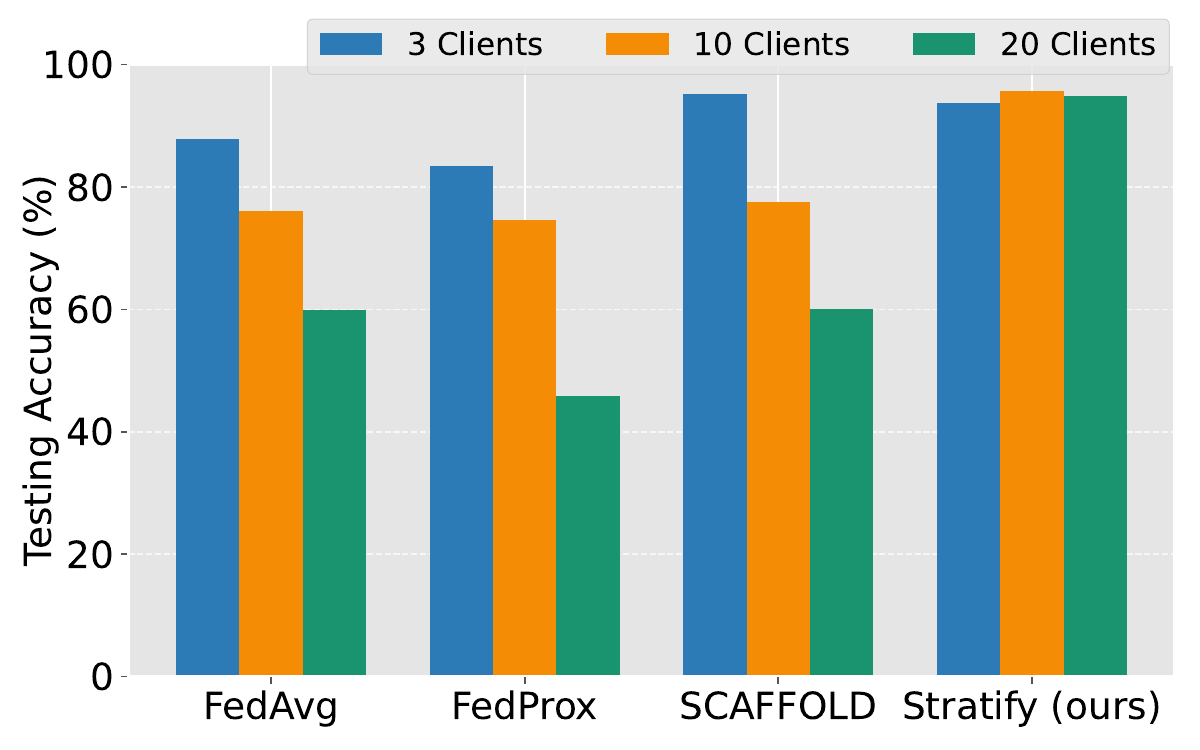}%
    \label{fig:chart12}}
    \hspace{-10pt}
    
    \caption{Comparison of algorithms' performance with increasing client numbers on different datasets and data partitions in batch-data learning}
    \label{fig:bar_charts2}
\end{figure*}

Fig. \ref{fig:bar_charts} and \ref{fig:bar_charts2} illustrate the effect of increasing client number on algorithms performance across different datasets and data partitions. In both single-sample and batch-data learning settings, FedAvg, FedProx, and SCAFFOLD generally exhibit a decline in accuracy as more clients participate, with some fluctuations in certain cases. The accuracy decline is due to the reduced local dataset size in the class and domain based partitions which constrain per-client learning, and the growing data imbalance in Dirichlet partition which further exacerbating the disparity between local updates. 
An opposite trend is observed for SFL. Notably, in Dirichlet partition, SFL achieves accuracy comparable to \shortname\ as more clients participate. This is because the data distribution becomes more spread out across clients as the client number increases, and reduces the risk of SFL overfitting due to prolonged training on a single client large data. However, SFL accuracies in the class and domain based partitions are still lower than \shortname\ due to catastrophic forgetting caused by disjoint client labels. 

All and all, \shortname\ maintains stable performance regardless of increasing client numbers because its global model update mechanisms are designed to be resilient to varying clients dataset sizes and imbalances. Instead of requiring each client to train on its entire dataset at once, \shortname\ manages the varying number of clients as a dynamic pool of available clients for each placeholder, in which the server can strategically select clients to perform local training on the few masked label that a training task requires each time. In this way, \shortname\ ensures that the contribution of client with small data is meaningfully integrated by accumulating its summed gradients as part of the contribution for a batch or directly incorporates its update into the single-sample learning model then only passing the model to the next available client.

\subsection{FedTAN Vs. \shortname\ Performance with BN} 

\begin{figure}[!t]
\centering
\includegraphics[width=6.0cm, height=4.5cm]{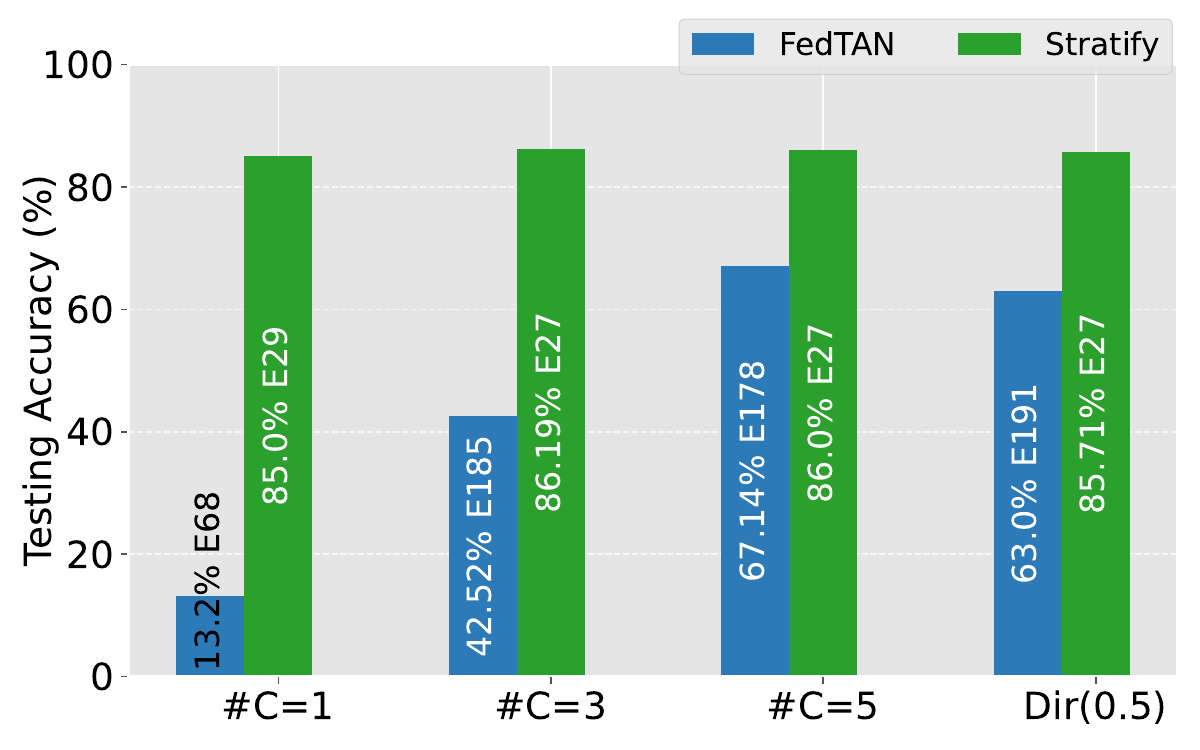}
\caption{FedTAN and \shortname\ performance on CIFAR-10 across various non-IID settings}
\label{fedtanvsfedgtc}
\end{figure}

The performance comparison between FedTAN and \shortname\ is separately conducted using ResNet-20 on the CIFAR-10 dataset, as FedTAN's custom BN code is only available for this model architecture. Its implementation heavily relies on numerous local arrays passing and computations, making it challenging for us to decouple and adapt to other model architectures. In Figure \ref{fedtanvsfedgtc}, we can observe that FedTAN struggles to achieve high accuracy across various non-IID settings and converges significantly slower than \shortname. This result indicates that merely addressing the batch statistics deviation is insufficient in a non-IID setting as it does not mitigate the underlying data distribution shift across clients. In contrast, since \shortname\ has the mechanisms that effectively prevent the adverse effects of non-IID data, the use of our custom BN layer in \shortname\ does not suffer from the same convergence issue observed in FedTAN. 

\begin{table}[t]
    \centering
    \caption{Total Training Time + Parameters/Gradients Serialization Deserialization and Weight Loading Time Per Round (Second) on CIFAR-10 with 10 Clients}
    \begin{tabular}{c@{}c|c|c}
        \toprule
        \textbf{Algorithm} & \textbf{\shortstack{Single Sample \\ Learning}} & \textbf{\shortstack{Batch Learning \\ (No BN)}} & \textbf{\shortstack{Batch Learning \\ (With BN)}} \\
        \midrule
        FedAvg & 53s + $\sim$0.006s & 11.4s + $\sim$0.127s & 11.5s + $\sim$0.154s \\
        FedProx & 53.4s + $\sim$0.006s & 9.83s + $\sim$0.127s & 10.1s + $\sim$0.154s \\
        Scaffold & 54.6s + $\sim$0.006s & 10.43s + $\sim$0.127s & 10.8s + $\sim$0.154s \\
        SFL & 142s + $\sim$0.06s & - & - \\
        FedTAN & - & - & 56s + $\sim$0.154s \\
        \shortname\ (Ours) & 145s + $\sim$95.4s & 186s + $\sim$11.8s & 249s + $\sim$12s \\
        \bottomrule
    \end{tabular}
    \label{tab:training_time}
\end{table}

\begin{table}[t]
    \centering
    \caption{Time (Second/Millisecond) to Construct Unique Real Labels List + Clients Label to Placeholder Mapping + Global Label Counts using Homomorphic Encryption. Each client holds \#C label of 1, 5, 50 for each respective total classes.}
    \footnotesize 
    \begin{tabular}{@{}c@{}c@{\hspace{2pt}}|c@{\hspace{2pt}}|c@{}}
        \toprule
        & \multicolumn{3}{c}{\textbf{Number of Clients}} \\
        \cmidrule(lr){2-4}
        \multicolumn{1}{c}{\textbf{Total Class}} & \textbf{5 Clients} & \textbf{10 Clients} & \textbf{20 Clients} \\
        \midrule
        2 Classes & 20{\scriptsize ms} + 0.8{\scriptsize s} + 8{\scriptsize ms} & 20{\scriptsize ms} + 0.8{\scriptsize s} + 9{\scriptsize ms} & 9{\scriptsize ms} + 0.9{\scriptsize s} + 10{\scriptsize ms} \\
        10 Classes & 0.4{\scriptsize s} + 1.3{\scriptsize s} + 40{\scriptsize ms} & 0.5{\scriptsize s} + 1.7{\scriptsize s} + 40{\scriptsize ms} & 0.6{\scriptsize s} + 2{\scriptsize s} + 50{\scriptsize ms} \\
        100 Classes & 63{\scriptsize s} + 19{\scriptsize s} + 0.4{\scriptsize s} & 90{\scriptsize s} + 22{\scriptsize s} + 0.5{\scriptsize s} & 102{\scriptsize s} + 27{\scriptsize s} + 0.6{\scriptsize s} \\
        \bottomrule
    \end{tabular}
    \label{tab:he_time}

\end{table}
\subsection{Algorithm Efficiency and Practical Deployment}

\textbf{Training Overhead:} 
As summarized in Table \ref{tab:training_time}, while \shortname\ learning approach increases the total training time and communication overhead per round due to its iterative, fine-grained learning using SLS. The number of rounds required for convergence using \shortname\ is markedly lower, with competitive local training workload per client (e.g., 14.5, 18.6, and 24.9 seconds on average per round) compared to conventional methods. This results in substantially lower overall local training time required per client across all rounds using \shortname. For instance, on the CIFAR-10 dataset, \shortname\ converges in just 8 rounds with an average overall local training time of 199 seconds per client\textemdash compared to FedAvg, which took 50 rounds, with a higher cumulative local training time of 575 seconds per client. The local training workload on each client further amplifies with the use of multiple local epochs in conventional methods. This highlights the efficiency and practicality of our learning approach, which achieve strong model performance with minimal per-client training effort. 

\textbf{Secure Client Selection with Masked Label Representation:} Our secure client selection protocol employs a masked label representation mechanism enabled by the CKKS homomorphic encryption scheme. 
Experimental evaluation (see Table \ref{tab:he_time}) indicates that the overhead introduced by this privacy‑preserving measure is modest and does not hinder the overall convergence of \shortname. 
The time to construct encrypted unique label list increases as the total number of classes grows, as more label vectors need to be processed by server. However, this is a one-time process. Once the unique label list reaches the required total class number, the server stops processing label vectors from any remaining clients to avoid redundant computations. The time for mapping encrypted real labels to placeholders for each client remains similar across different client numbers with concurrent processing.

\subsection{Ablation Studies and Further Analysis}
\label{ablationStudy}
We conducted non-IID experiments on CIFAR-10 using Dirichlet-based label distribution, and on the PACS and Digit-DG datasets with one of the domains concentrated on a single client. These setups are used to evaluate \shortname\ performance under both uniform and weighted client selection strategies, and to examine the effect of varying chunk size in the single-sample learning setting.

As shown in Table \ref{tab:uniform_vs_weighted}, in CIFAR-10, \shortname\ performs equally well with both uniform and weighted client selection, as the effects of skewed label distributions are mitigated with the stratified labels in SLS. Chunk size 5 results in around 35\% reduction in model transfer frequency, compared to naive random passing for each label in SLS. Increasing the chunk size provides only an additional few percents reduction. In the non-IID setup for PACS and Digit-DG, uniform client selection leads to training being dominated in later training iterations by this single client that holds a large single domain data. Despite this, the single-sample learning setting demonstrates resilience, achieving accuracies comparable to weighted selection, especially with a smaller chunk size. The batch learning setting, on the other hand, shows slower convergence with uniform selection, but still able to achieve accuracy close to weighted selection. Hence, our experimental results and ablation study confirm that the secure uniform client selection is effective for most non-IID scenarios, highlighting the robustness and practicality of our approach in handling a variety of data distributions, when stringent label security is required.

\begin{table}[t]
    \centering
    \caption{Testing Accuracy Obtained Using Uniform (U) and Weighted (W) Client Selection for 10 Clients. E\# denotes the epoch \shortname\ achieves the highest accuracy, CS\# denotes the placeholder chunk size used for single-sample selection}
    \footnotesize 
    \begin{tabular}{c|c|c|c}
        \toprule
        \textbf{Setting} & \textbf{CIFAR-10} & \textbf{PACS} & \textbf{Digit-DG} \\
        \midrule
        Single-Sample, U, CS5 & 80.4\%, E28 & 95.2\%, E15 & 82.5\%, E13 \\
        Single-Sample, U, CS50 & 80.2\%, E29 & 94.9\%, E15 & 81.5\%, E10 \\
        Single-Sample, U, CS150 & 80.1\%, E26 & 94.9\%, E15 & 80.8\%, E15 \\
        Single-Sample, W, CS5 & 80.2\%, E30 & 95.4\%, E11 & 81.1\%, E10 \\
        Single-Sample, W, CS50 & 80.7\%, E26 & 95.5\%, E15 & 81.3\%, E14 \\
        Single-Sample, W, CS150 & 80.1\%, E26 & 95.4\%, E14 & 81.1\%, E14 \\ 
        Batch-Data, U & 90.4\%, E8 & \textbf{93.7\%, E18} & \textbf{84.6\%, E11} \\ 
        Batch-Data, W & 90.3\%, E8 & 94.4\%, E9 & 85.2\%, E9 \\ 
        \bottomrule
    \end{tabular}
    \label{tab:uniform_vs_weighted}
\end{table}

\begin{table}[t]
    \centering
    \caption{Testing Accuracy Obtained Using ResNet-9 with Group Normalization (GN) layer for 10 Clients.\\ E\# denotes the epoch \shortname\ achieves the highest accuracy.}
    \footnotesize
    \begin{tabular}{c|c|c}
        \toprule
        \textbf{Setting} & \textbf{CIFAR-10} & \textbf{CIFAR-100} \\
        \midrule
        Single-Sample Learning & 85.1\%, E8 & 61.9\%, E30 \\
        Batch-Data Learning & 83\%, E8 & 66.1\%, E38 \\
        \bottomrule
    \end{tabular}
    \label{tab:gn_acc}
\end{table}

In addition, we further examined the single-sample learning setting using a more complex model architecture: ResNet-9 with Group Normalization (GN). This experiment aimed to confirm whether the single-sample learning, previously evaluated with simpler CNN models in the main experiments, could also deliver strong performance on a deeper architecture. As shown in Table \ref{tab:gn_acc}, ResNet-9 with GN outperformed the simpler CNN models on CIFAR-10 and CIFAR-100 in the single-sample learning setting, highlighting the adaptability of the learning setting to various model architectures.

\subsection{Discussion Summary}
The experimental results provide compelling evidence that \shortname\ effectively addresses the challenges posed by non‑IID data in federated learning. By integrating the SLS, secure label‑aware client selection, and novel single‑sample and batch-data learning, \shortname\ achieves robust global updates, rapid convergence, and high accuracy under diverse conditions. Moreover, our communication efficiency analysis demonstrates that the additional privacy‑preserving measures and frequent updates are balanced by a reduction in the total number of training rounds. This evidence strongly supports that \shortname\ offers a practical and effective solution to the inherent challenges of non‑IID federated learning.

\section{Conclusion}
In this work, we demonstrate that addressing the design limitations of Fedvg through our novel guided training mechanism, enables global model to effectively learn across various non-IID settings. While \shortname\ incurs high communication overhead due to frequent model updates, it is able to consistently maintain and achieve high testing accuracy across all non-IID scenarios with lower training rounds compared to all baselines. We suggest future work to focus on improving the efficiency of \shortname\ or explore other adaptive global learning rebalancing methods that can effectively handle non-IID data while ensuring computational efficiency. Since we primarily introduce novel training approach for effective robust model training in non-IID settings, we leave security challenges such as data poisoning and model inversion attacks as another key future work to address. These advancements could further enhance the real-world applicability of the proposed approach, and pave the way for more effective, efficient, and secure federated learning in non-IID settings.


 

\bibliographystyle{IEEEtran} 
\bibliography{references}

\begin{thebibliography}{10}
\providecommand{\url}[1]{#1}
\csname url@samestyle\endcsname
\providecommand{\newblock}{\relax}
\providecommand{\bibinfo}[2]{#2}
\providecommand{\BIBentrySTDinterwordspacing}{\spaceskip=0pt\relax}
\providecommand{\BIBentryALTinterwordstretchfactor}{4}
\providecommand{\BIBentryALTinterwordspacing}{\spaceskip=\fontdimen2\font plus
\BIBentryALTinterwordstretchfactor\fontdimen3\font minus \fontdimen4\font\relax}
\providecommand{\BIBforeignlanguage}[2]{{%
\expandafter\ifx\csname l@#1\endcsname\relax
\typeout{** WARNING: IEEEtran.bst: No hyphenation pattern has been}%
\typeout{** loaded for the language `#1'. Using the pattern for}%
\typeout{** the default language instead.}%
\else
\language=\csname l@#1\endcsname
\fi
#2}}
\providecommand{\BIBdecl}{\relax}
\BIBdecl

\bibitem{guerra2023cost}
E.~Guerra, F.~Wilhelmi, M.~Miozzo, and P.~Dini, ``The cost of training machine learning models over distributed data sources,'' \emph{IEEE Open Journal of the Communications Society}, vol.~4, pp. 1111--1126, 2023.

\bibitem{Briggs2020}
C.~Briggs, Z.~Fan, and P.~Andras, ``{Federated learning with hierarchical clustering of local updates to improve training on non-IID data},'' in \emph{2020 International Joint Conference on Neural Networks (IJCNN)}.\hskip 1em plus 0.5em minus 0.4em\relax IEEE, 2020, pp. 1--9.

\bibitem{jothimurugesan2023federated}
E.~Jothimurugesan, K.~Hsieh, J.~Wang, G.~Joshi, and P.~B. Gibbons, ``Federated learning under distributed concept drift,'' in \emph{International Conference on Artificial Intelligence and Statistics}.\hskip 1em plus 0.5em minus 0.4em\relax PMLR, 2023, pp. 5834--5853.

\bibitem{Vahidian2023}
S.~Vahidian, M.~Morafah, W.~Wang, V.~Kungurtsev, C.~Chen, M.~Shah, and B.~Lin, ``{Efficient distribution similarity identification in clustered federated learning via principal angles between client data subspaces},'' in \emph{Proceedings of the AAAI Conference on Artificial Intelligence}, vol.~37, no.~8, 2023, pp. 10\,043--10\,052.

\bibitem{Xiao2021}
Y.~Xiao, J.~Shu, X.~Jia, and H.~Huang, ``{Clustered federated multi-task learning with non-IID data},'' in \emph{2021 IEEE 27th International Conference on Parallel and Distributed Systems (ICPADS)}.\hskip 1em plus 0.5em minus 0.4em\relax IEEE, 2021, pp. 50--57.

\bibitem{Huang2021}
Y.~Huang, L.~Chu, Z.~Zhou, L.~Wang, J.~Liu, J.~Pei, and Y.~Zhang, ``{Personalized cross-silo federated learning on non-iid data},'' in \emph{Proceedings of the AAAI conference on artificial intelligence}, vol.~35, no.~9, 2021, pp. 7865--7873.

\bibitem{Luo2022}
J.~Luo and S.~Wu, ``{Adapt to adaptation: Learning personalization for cross-silo federated learning},'' in \emph{IJCAI: proceedings of the conference}, vol. 2022.\hskip 1em plus 0.5em minus 0.4em\relax NIH Public Access, 2022, p. 2166.

\bibitem{Chen2023}
H.~Chen, A.~Frikha, D.~Krompass, J.~Gu, and V.~Tresp, ``{FRAug: Tackling federated learning with Non-IID features via representation augmentation},'' in \emph{Proceedings of the IEEE/CVF International Conference on Computer Vision}, 2023, pp. 4849--4859.

\bibitem{li2024feature}
Z.~Li, Y.~Sun, J.~Shao, Y.~Mao, J.~H. Wang, and J.~Zhang, ``Feature matching data synthesis for non-iid federated learning,'' \emph{IEEE Transactions on Mobile Computing}, vol.~23, no.~10, pp. 9352--9367, 2024.

\bibitem{Li2020}
T.~Li, A.~K. Sahu, M.~Zaheer, M.~Sanjabi, A.~Talwalkar, and V.~Smith, ``{Federated optimization in heterogeneous networks},'' \emph{Proceedings of Machine learning and systems}, vol.~2, pp. 429--450, 2020.

\bibitem{Xu2022}
A.~Xu and H.~Huang, ``{Coordinating momenta for cross-silo federated learning},'' in \emph{Proceedings of the AAAI Conference on Artificial Intelligence}, vol.~36, no.~8, 2022, pp. 8735--8743.

\bibitem{Karimireddy2020}
S.~P. Karimireddy, S.~Kale, M.~Mohri, S.~Reddi, S.~Stich, and A.~T. Suresh, ``{Scaffold: Stochastic controlled averaging for federated learning},'' in \emph{International conference on machine learning}.\hskip 1em plus 0.5em minus 0.4em\relax PMLR, 2020, pp. 5132--5143.

\bibitem{zhao2023ensemble}
Z.~Zhao, J.~Wang, W.~Hong, T.~Q. Quek, Z.~Ding, and M.~Peng, ``Ensemble federated learning with non-iid data in wireless networks,'' \emph{IEEE Transactions on Wireless Communications}, vol.~23, no.~4, pp. 3557--3571, 2023.

\bibitem{Shi2023}
N.~Shi, F.~Lai, R.~{Al Kontar}, and M.~Chowdhury, ``{Fed-ensemble: Ensemble models in federated learning for improved generalization and uncertainty quantification},'' \emph{IEEE Transactions on Automation Science and Engineering}, 2023.

\bibitem{mcmahan2017communication}
B.~McMahan, E.~Moore, D.~Ramage, S.~Hampson, and B.~A. y~Arcas, ``Communication-efficient learning of deep networks from decentralized data,'' in \emph{Artificial intelligence and statistics}.\hskip 1em plus 0.5em minus 0.4em\relax PMLR, 2017, pp. 1273--1282.

\bibitem{thompson2012sampling}
S.~K. Thompson, \emph{Sampling}.\hskip 1em plus 0.5em minus 0.4em\relax John Wiley \& Sons, 2012, vol. 755.

\bibitem{hamilton2020time}
J.~D. Hamilton, \emph{Time series analysis}.\hskip 1em plus 0.5em minus 0.4em\relax Princeton university press, 2020.

\bibitem{Bao2023}
W.~Bao, H.~Wang, J.~Wu, and J.~He, ``{Optimizing the collaboration structure in cross-silo federated learning},'' in \emph{International Conference on Machine Learning}.\hskip 1em plus 0.5em minus 0.4em\relax PMLR, 2023, pp. 1718--1736.

\bibitem{fraboni2021clustered}
Y.~Fraboni, R.~Vidal, L.~Kameni, and M.~Lorenzi, ``Clustered sampling: Low-variance and improved representativity for clients selection in federated learning,'' in \emph{International Conference on Machine Learning}.\hskip 1em plus 0.5em minus 0.4em\relax PMLR, 2021, pp. 3407--3416.

\bibitem{gao2024fedsts}
D.~Gao, D.~Song, G.~Shen, X.~Cai, L.~Yang, G.~Liu, X.~Li, and Z.~Wang, ``Fedsts: A stratified client selection framework for consistently fast federated learning,'' \emph{IEEE Transactions on Neural Networks and Learning Systems}, 2024.

\bibitem{li2023convergence}
Y.~Li and X.~Lyu, ``Convergence analysis of sequential federated learning on heterogeneous data,'' \emph{Advances in Neural Information Processing Systems}, vol.~36, pp. 56\,700--56\,755, 2023.

\bibitem{Wang2023}
Y.~Wang, Q.~Shi, and T.-H. Chang, ``{Why batch normalization damage federated learning on non-iid data?}'' \emph{IEEE transactions on neural networks and learning systems}, 2023.

\bibitem{li2022federated}
Q.~Li, Y.~Diao, Q.~Chen, and B.~He, ``Federated learning on non-iid data silos: An experimental study,'' in \emph{2022 IEEE 38th international conference on data engineering (ICDE)}.\hskip 1em plus 0.5em minus 0.4em\relax IEEE, 2022, pp. 965--978.

\end{thebibliography}




\end{document}